\documentclass[preprint,12pt]{elsarticle}

\usepackage{algorithm}
\usepackage{algpseudocode}
\usepackage{multirow}
\usepackage{url}
\usepackage{xcolor}
\usepackage{amssymb}
\usepackage{amsmath}
\usepackage{amsthm}
\newtheorem{proposition}{Proposition}
\journal{Nuclear Physics B}

\begin{document}

\begin{frontmatter}

\title{PHISHREV+: An ASP-Based Post-Hoc 
Defeasible Machine Learning Framework for Phishing Detection}

\author[label1,label3]{Mainak Sen}
\author[label1,label2]{Kumar Sankar Ray}
\author[label3,label4]{and Amlan Chakrabarti}

\affiliation[label1]
{organization={Computer Science \& Engineering Department, Techno India University, West Bengal},
            addressline={EM 4, Salt Lake}, 
            city={Kolkata},
            postcode={700091}, 
            state={West Bengal},
            country={India}}
\affiliation[label2]{organization={ECSU Unit, Indian Statistical Institute},
            addressline={203, B. T Road}, 
            city={Kolkata},
            postcode={700108}, 
            state={West Bengal},
            country={India}}

\affiliation[label3]{organization={A.K.Choudhury School of Information Technology, University of Calcutta},
            addressline={JD 2, Sector 3}, 
            city={Kolkata},
            postcode={700106}, 
            state={West Bengal},
            country={India}}            

\affiliation[label4]{organization={Department of Artificial Intelligence, Indian Institute of Technology},
            addressline={Kharagpur}, 
            city={Kolkata},
            postcode={721302}, 
            state={West Bengal},
            country={India}}


\begin{abstract}
Phishing website detection remains a critical cybersecurity challenge. Although machine learning classifiers achieve high accuracy on benchmark datasets, they suffer from a fundamental limitation: once trained, their decision boundaries are fixed and cannot incorporate new contextual evidence without retraining the model. This work presents \textbf{PHISHREV+}, a hybrid framework that instantiates the notion of \textit{post-hoc defeasible machine learning}. This framework does not replace the classifier. It treats the initial classifier predictions as defeasible beliefs and revises them through an Answer Set Programming based non-monotonic reasoning(NMR) layer. The framework operates in two phases. In the machine learning phase, four classifiers (SVM, KNN, DT and RF) are trained on the 87 feature benchmark dataset. In the post-hoc NMR phase, four defeasible rules derived from exploratory data analysis of the training set, and validated through Chi-square and Mann-Whitney U tests, formally revise initial predictions. The proposed framework achieves consistent false positive reduction across all our classifiers SVM (41$\rightarrow$39), KNN 
(47$\rightarrow$46), DT 
(62$\rightarrow$61), and RF 
(52$\rightarrow$51) while maintaining complete false negative parity with zero increase overhead. A comprehensive ablation studies across seven rule configurations confirms that the hierarchical combination of all four defeasible rules is necessary to achieve this optimal balance. Instance level explainability is provided through eXplainable ASP (XASP), which generates formal derivation graphs showing which defeasible rules fired and why the initial belief was retracted, a capability absent from all existing phishing detection approaches.

\end{abstract}


\end{frontmatter}



\section{Introduction}
\label{sec1}
\begin{figure*}[htbp]
    \centering
    \includegraphics[width=14cm, height=9cm]{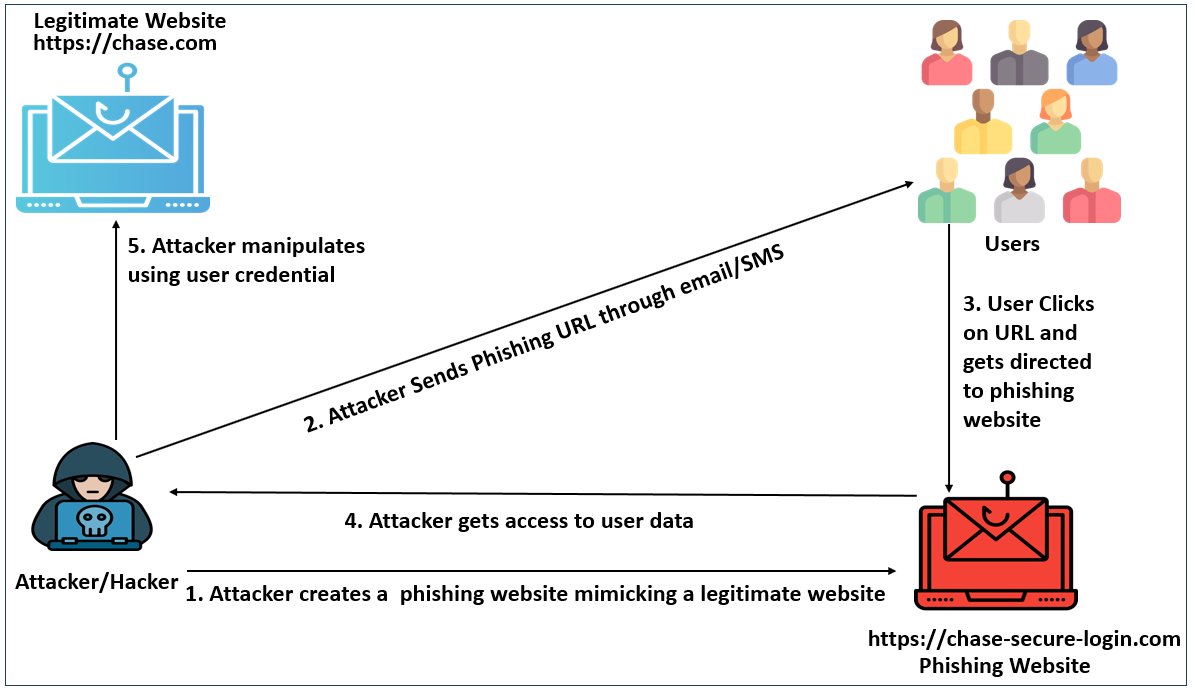}
    \caption{Anatomy of a phishing attack. The attacker creates a look-alike
    site (\texttt{https://chase-secure-login.com}) mimicking the legitimate
    domain (\texttt{https://chase.com}) and distributes its URL via email/SMS.
    Victim data flows from the user to the phishing site and then to the
    attacker, who reuses the stolen credentials on the genuine site.}
    \label{fig:phishing_attack}
\end{figure*}

The world wide web(WWW) has transformed how individuals, organizations, government communicate and share information. The web operates through Uniform Resource Locators(URL), a structured string that identify the location of resources on the internet. A URL typically comprises a protocol identifier, a domain name, optional subdomains, a path and query parameters. As web pages has grown significantly, the sophistication of web based threats has also increased that exploit this web system for malicious purposes. URLs serve as the entry point for many cyber-attacks such as phishing\cite{zieni2023phishing}, drive-by-download\cite{nappa2015malicia}. \\
Among web-based threats, phishing remains one of the most pervasive and damaging \cite{mosa2023machine,asif2023machine}. A phishing webpage is a fraudulent website crafted to impersonate a legitimate website. The motive of this attack is to deceive human to disclose sensitive and personal credentials. Figure\ref{fig:phishing_attack} illustrates the phishing attack life cycle. According to the Anti-Phishing Working Group (APWG)\cite{apwg2026q2}, 1,069,681 phishing attacks were observed in the second quarter of 2026, an increase of 10.1\% over the previous quarter 2026(January-March). At this volume, even a small false-positive rate produces a large number of wrong alerts, which adds to alert fatigue of security analysts\cite{ban2023breaking}. The impact of phishing attack goes beyond financial loses, often resulting to data breaches to reputational damage \cite{ putra2024analysis}. Unlike malware or network intrusions, phishing exploits human psychology rather than technical vulnerabilities, making it difficult to defend against using conventional security mechanism. Automated detection of phishing URLs is therefore a critical requirement for modern cybersecurity infrastructure. 

Significant research effort has been made toward automated phishing detection. Early approaches relied on blacklist and heuristic rules\cite{prakash2010phishnet, mohammad2014intelligent, da2020heuristic, barraclough2021intelligent}but these suffer from high maintenance overhead and inability to detect unseen phishing pages. Machine learning has emerged as a dominant paradigm with classifiers trained on lexical, structural and content based features achieving high accuracy\cite{odeh2021machine,ahammad2022phishing, jalil2023highly,mosa2023machine,kocyigit2024enhanced,kalla2023phishing}. More recently, deep learning approaches including convolutional neural network \cite{le2018urlnet}, recurrent neural network architectures and hybrid architectures \cite{wang2019pdrcnn, akccam2025development}, transformer based architectures \cite{aljofey2025bert}have been applied to phishing detection. Their ability to learn hierarchical feature representations directly from raw URL strings and webpage content eliminates the need for extensive feature engineering. Large language models have also been explored for phishing detection, exploiting their pre-trained linguistic knowledge to identify deceptive content \cite{li2025phishdebate, roy2024phishlang}. They achieved rapid inference but freezing a decision boundary at training time can only be updated through wholesale retraining. Multimodal systems \cite{lokesh2026multimodal,bao2025phishsense} have further extends to combine URL, content and visual signals, implicitly acknowledging that no single fixed is sufficient. \\
However, machine learning, deep learning and LLM-based approaches share a common limitation that once a classifier is trained or established, their decision boundaries are fixed lacking the ability of a principled defeasibility mechanism. Incorporating new domain knowledge such as emerging phishing tactics or newly identified adversarial signals, requires full model retraining. For random forest classifier, this involves a complexity of $\mathcal{O}(T \cdot n \cdot f \cdot \log n)$, while deep learning and LLM based fine tuning incur higher computational cost due to gradient based optimization over millions of parameters. This retraining of trained classifiers is computationally expensive as well as operationally disruptive, particularly in real time security environment where rapid responses to new threat is needful. Furthermore, phishing strategies evolve continuously over time, causing trained models to degrade in performance due to concept drift\cite{li2024concept, ejaz2023life}. Features that reliably distinguished phishing from legitimate URLs in earlier datasets may lose their discriminative power over time and adversaries adapt their tactics. As statistical classifiers cannot incorporate contextual reasoning, they cannot revise a prediction when new and contradictory evidence becomes available, leading to persistent false positives that erode user trust and contribute to alert fatigue\cite{ban2023breaking}.     \\
The objective of this work is to address these limitations by proposing a hybrid framework that combines the pattern recognition ability of machine learning with the contextual reasoning ability of non-monotonic reasoning. Specifically, we aim to enable post-hoc belief revision of initial classifier predictions using EDA-derived contextual evidence, without retraining the underlying model. The 
following research questions guide 
this work:
\begin{itemize}

\item \textbf{RQ1:} To what extent can exploratory data analysis(EDA) of contextual URL and webpage features identify statistically  significant distributional differences between phishing and legitimate websites, thereby providing evidence for the formulation of defeasible revision rules?

\item \textbf{RQ2:} TO what extent can Answer Set Programming (ASP)-based non-monotonic reasoning (NMR) use EDA-derived contextual evidence to revise the initial predictions of machine learning classifiers, reducing false positives while preserving false-negative performance? 

\item \textbf{RQ3:} 
How does hierarchical interaction of multiple defeasible rules influence the trade-off between false-positive reduction and false-negative preservation? Furthermore, what advantages does an ASP-based formulation offer over equivalent deterministic post-processing approaches in terms of rule interaction, conflict resolution and explainability?

\item \textbf{RQ4:} Can the proposed reasoning layer incorporate newly available domain knowledge through rule updates in  $\mathcal{O}$ time, without retraining the underlying machine learning classifiers and thereby support efficient post-hoc adaption to evolving phishing contexts?

\end{itemize}

To address the research questions, we propose \textbf{PHISHREV+}, a dual-stage framework comprising a machine learning phase and a post-hoc ASP-based non-monotonic reasoning phase. In the first phase, four classifiers namely Support Vector Machine(SVM), K-Nearest Neighbour(KNN), Decision Tree(DT) and Random Forest(RF) are trained on the 87-feature benchmark dataset proposed by \cite{hannousse2021web}. In the second phase an EDA analysis is conducted on four contextual features \textit{meta\_info\_present}, \textit{length\_url}, \textit{nb\_www}, and \textit{domain\_in\_title} to derive four hierarchical defeasible rules encoded within ASP knowledge base. The NMR layer accepts initial classifier predictions as beliefs and revises them using these rules via the Clingo ASP solver. Experimental evaluation demonstrates consistent false positive reduction with zero false negative increase parity with the base initial classifier. The novel contributions of this work are as follows:
\begin{itemize}

\item  An exploratory data analysis of four contextual features \textit{meta\_info\_present}, \textit{length\_url}, \textit{nb\_www}, 
and \textit{domain\_in\_title} is conducted to identify statistically meaningful distributional differences between phishing and legitimate URLs. The feature \textit{nb\_www} emerges as the strongest binary signal with a separation of 52.8 percentage points. The feature \textit{domain\_in\_title} shows a counter-intuitive pattern. In the subset that reaches rule 4, 82.6\% of phishing URLs embed the domain in the page title against 56.8\% of legitimate URLs.  
\item A 4-rule ASP-based non-monotonic reasoning framework is proposed that performs post-hoc belief revision of initial classifier predictions using hierarchical defeasible rules derived from EDA. This achieved consistent false positive reduction across all four classifier while maintaining complete false negative parity with the base classifier. 

\item A comprehensive ablation study across seven rule configurations shows that only the hierarchical combination of all four defeasible rules achieves false positive reduction with zero false negative increase.  A deterministic if-else baseline yields identical predictions, positioning the value of the ASP formulation in its declarative representation, formal semantics and support for rule-level explanation rather than its predictive performance.  

\item The reasoning layer supports the incorporation of new domain knowledge in  $\mathcal{O}(n)$ time without retraining the underlying classifiers, as established by a complexity analysis and confirmed empirically by the linear scaling of grounding and solving time and by the incremental rule additions of the ablation study.  

\end{itemize}
The remainder of the paper is as follows. Section \ref{survey} reviews the related work on phishing detection. The next section ~\ref{nmrintroduction} introduces non-monotonic reasoning, answer set programming and defeasible machine learning. Then section \ref{datasetinfo} states the dataset and features used for the experimentation. Next, exploratory data analysis of contextual features is presented in section \ref{eda}.
Section \ref{sec:framework} then describes the proposed dual framework and the 4-rule NMR knowledge base construction. Then experimental results with ablation study is shown in section \ref{result} and \ref{sec:ablation} respectively. Section \ref{stateofartworks} compares PHISHREV+ with prior work and XASP-based explainability is presented in section \ref{sec:xasp}. Section \ref{discussion} discusses the findings and limitations. Finally, in section \ref{conclusion}, conclusion of the paper with future direction is presented.  

\section{Related Work}\label{survey}
This section presents existing approaches to phishing detection, including machine learning, deep learning methods and recent advances in hybrid frameworks. 
\subsection{Blacklist-Based Approaches}
Security protocols can be categorized into software based approaches namely blacklist and whitelist \cite{das2019sok} based. Blacklist contains those which user should not visit while white-list contains all legitimate entries. When a user attempts to enter a URL, it is checked against the blacklist and blocked is a match is found. \\
Blacklist based systems are simple to implement and computationally inexpensive at query time. However, they suffer from various limitations. Firstly, these are inherently reactive, as a URL can only be added to blacklist after it has been identified. But, newly created phishing websites remain undetected during the interval between their deployment and inclusion in the blacklist. Secondly, attackers employ various 
evasion, obfuscation techniques \cite{skula2024url}, allowing malicious websites to bypass static blacklist mechanism. Thirdly, this approach requires continuous updates and maintenance to remain updated. Finally, blacklist does not provide any generalization capability. \\
These limitations motivated the transition toward machine learning based approaches that offer the ability to generalise from learned patterns and detect previously unseen phishing webpages.  

\subsection{Machine Learning Based Approaches}

Machine learning approaches extract discriminative features from URL and webpage content, which are subsequently used to train classification models for identifying phishing websites. Traditional ml algorithms like Random Forest (RF), Naïve Bayes (NB), Support Vector Machine(SVM), k-Nearest Neighbor (kNN) have been widely adopted \cite{odeh2021machine} for phishing URL classification and have demonstrated promising performances. Table \ref{tab:ml_survey} presents the summerized version of machine learning based approaches.

\begin{table*}[h]
\centering
\caption{Summary of Machine Learning Based 
Phishing Detection Approaches}
\label{tab:ml_survey}
\small
\begin{tabular}{|p{.6cm}|p{0.68cm}|p{2.2cm}|p{1.8cm}|p{4.5cm}|p{3.0cm}|}
\hline
\textbf{Ref.} & 
\textbf{Year} & 
\textbf{Algorithm} & 
\textbf{Features} & 
\textbf{Key Findings} & 
\textbf{Limitations} \\
\hline
\cite{hannousse2021towards} & 
2021 & RF & 
87 lexical, content and external & 
RF achieved 96.61\% accuracy on proposed 
87-feature benchmark dataset & 
No defeasibility; retraining required \\
\hline
\cite{gupta2021novel} & 
2021 & RF & 
9 lexical & 
RF achieved 99.57\% for real-time detection 
on constrained devices & 
No defeasibility; retraining required \\
\hline
\cite{adane2023single} & 
2023 & RF & 
87 lexical & 
RF achieved 96.54\% on benchmark dataset & 
No defeasibility; retraining required \\
\hline
\cite{choudhary2023machine} & 
2023 & RF, LR, DT, SVM, XGBoost & 
25 hybrid & 
RF achieved 98.80\% on PhishTank and 
97.87\% on UCI dataset & 
No defeasibility; retraining required \\
\hline
\cite{shafin2025explainable} & 
2025 & RF + XAI & 
87 lexical & 
RF achieved 96.76\% with SHAP-based 
explainability & 
No defeasibility; retraining required \\
\hline
\cite{uddin2025explainable} & 
2025 & XGBoost, GBM, LGBM + SHAP & 
30 hybrid URL & 
XGBoost achieved 97.78\% with SHAP-based 
explainability & 
No defeasibility; retraining required \\
\hline

\end{tabular}
\end{table*}

\subsection{Deep Learning Based Approaches}
Various deep learning (DL)based architectures overcome the need of manual feature engineering as they can extract the features automatically from raw data. The convolutional neural network (CNN), deep neural network (DNN), recurrent neural network(RNN), Long Short-Term Memory (LSTM), Gated Recurrent Unit (GRU) and deep belief network (DBN) are frequently used architectures\cite{do2022deep} by researchers. Table \ref{tab:dl_survey} presents a summary of deep learning research work done on phishing detection. 

\begin{table*}[h]
\centering
\caption{Summary of Deep Learning Based 
Phishing Detection Approaches}
\label{tab:dl_survey}
\small
\begin{tabular}{|p{.6cm}|p{0.6cm}|p{2.5cm}|p{5.5cm}|p{3.0cm}|}
\hline
\textbf{Ref.} & 
\textbf{Year} & 
\textbf{Algorithm} & 
\textbf{Key Findings} & 
\textbf{Limitations} \\
\hline
\cite{wang2019pdrcnn} & 
2019 & 
Bi-LSTM + CNN & 
Bi-LSTM followed by CNN achieved 
95.6\% accuracy on 500,000 URLs dataset & 
No defeasibility; retraining required \\
\hline
\cite{afzal2021urldeepdetect} & 
2021 & 
LSTM & 
LSTM achieved 98.3\% accuracy on 
450,176 URLs dataset & 
No defeasibility; retraining required \\
\hline
\cite{xiao2021phishing} & 
2021 & 
CNN + Multi-head Attention & 
CNN with multi-head attention 
achieved 95.6\% accuracy & 
No defeasibility; retraining required \\
\hline
\cite{alshehri2022character} & 
2022 & 
CNN & 
Character-level CNN achieved 
98.13\% accuracy & 
No defeasibility; retraining required \\
\hline
\cite{asiri2024phishingrtds} & 
2024 & 
Bi-LSTM + Attention & 
Bi-LSTM with attention achieved 
99\% accuracy on webpage content & 
No defeasibility; retraining required \\
\hline
\cite{gupta2024optimized} & 
2024 & 
RNN + WOA & 
WOA-optimized RNN achieved 92\% 
accuracy on 11K dataset & 
No defeasibility; retraining required \\
\hline
\cite{akccam2025development} & 
2025 & 
Bi-LSTM + Attention & 
Bi-LSTM with attention achieved 
best accuracy of 98.34\% & 
No defeasibility; retraining required \\
\hline
\end{tabular}
\end{table*}

\subsection{Large Language Model Based Approaches}
Large Language models like GPT-4, Claude, and Llama have recently been explored for phishing detection. Table \ref{tab:llm_survey} summarizes representative LLM-based phishing based approaches.  
\begin{table*}[h]
\centering
\caption{Summary of LLM Based 
Phishing Detection Approaches}
\label{tab:llm_survey}
\small
\begin{tabular}{|p{.6cm}|p{0.7cm}|p{2.5cm}|p{5.5cm}|p{3.5cm}|}
\hline
\textbf{Ref.} & 
\textbf{Year} & 
\textbf{Model} & 
\textbf{Key Findings} & 
\textbf{Limitations} \\
\hline
\cite{heiding2024devising} & 
2024 & 
GPT-4 & 
GPT-4 used for both generating and 
detecting phishing emails. Claude 
detected AI-generated phishing with 
100\% accuracy when primed but dropped 
to 75\% on real-world emails & 
Non-revisable at inference; 
fine-tuning required for updates \\
\hline
\cite{chataut2024llm} & 
2024 & 
GPT-3.5, GPT-4, 
ChatGPT & 
Evaluated phishing detection 
capabilities of three LLM variants 
with varying levels of success 
across models & 
Non-revisable at inference; 
fine-tuning required for updates \\
\hline
\cite{zhang2025benchmarking} & 
2025 & 
GPT-3.5, Llama-3 & 
GPT-3.5 and Llama-3-8b-Instruct 
identified as best performing models 
for phishing detection in SMEs & 
Non-revisable at inference; 
fine-tuning required for updates \\
\hline
\cite{lee2024multimodal} & 
2024 & 
Multimodal LLM & 
Multimodal LLM combining webpage 
visual and textual signals for 
phishing webpage detection & 
Non-revisable at inference; 
fine-tuning required for updates \\
\hline
\end{tabular}
\end{table*}

\subsection{Research Gap}
The survey of existing phishing detection approaches, spanning blacklist-based, 
machine learning, deep learning, and large language model based methods, reveals a consistent and fundamental limitation that none of the existing approaches incorporate a principle defeasibility mechanism that enables post-hoc belief revision when new evidences become available at inference time. This gap motivates, the proposed framework, which addresses these limitations by integrating machine learning classifiers with an ASP-based non-monotonic reasoning layer for post-hoc belief revision. Unlike existing approaches, PHISHREV+ derives defeasible rules directly from exploratory data analysis of features, enabling selective and evidence-driven belief correction in $\mathcal{O}(n)$ time without modifying or retraining the underlying ML classifiers.  

\section{Background and Preliminaries}
\label{nmrintroduction}
In this section, we introduce the principle of non monotonic reasoning and its formal characterization 
through defeasible logic. We then formally contrast the decision making paradigm of machine learning classifier 
with that of NMR-based reasoning systems, highlighting the fundamental distinction between fixed rule statistical decision functions and defeasible belief revisions. 
\subsection{Non-monotonic Reasoning}
Classical deductive reasoning is monotonic in nature that is adding new premises to an existing knowledge base never invalidates previously derived conclusions. Formally saying, if a conclusion $\phi$ is derivable from 
a set of premises $\Gamma$, it remains derivable from any superset $\Gamma' \supseteq \Gamma$ as shown in equation \ref{eq:monotonic}.
\begin{equation}
\Gamma \vdash \phi \Rightarrow 
\Gamma' \vdash \phi, \quad 
\forall\ \Gamma' \supseteq \Gamma
\label{eq:monotonic}
\end{equation}
This property is in mathematical proof systems, it is ill-suited to real world reasoning scenarios where new evidences may contradict or override earlier conclusion. Non-monotonic reasoning addresses this limitation by allowing previously derived conclusions to be formally withdrawn upon the arrival of new and contradictory evidence\cite{brewka1997nonmonotonic} presented in equation \ref{eq:nonmonotonic}.

\begin{equation}
\Gamma \vdash \phi \quad \text{but} \quad 
\Gamma' \not\vdash \phi, \quad 
\text{for some}\ \Gamma' \supseteq \Gamma
\label{eq:nonmonotonic}
\end{equation}

This property known as \textit{defeasibility} which enables a reasoning system to revise its beliefs in a principled and auditable manner, without discarding the underlying knowledge base. A defeasible rule takes the general form as shown in equation \ref{eq:defeasible}
\begin{equation}
P \wedge \neg Q \rightarrow C
\label{eq:defeasible}
\end{equation}

where, $P$ presents the body of the rule, $C$ is the conclusion and $Q$ is a defeating condition whose absence is required for the conclusion to hold. If $Q$ becomes derivable from new evidence, the conclusion $C$ is formally retracted even if $P$ remains true. This is the nature of non-monotonic \cite{reiter1988nonmonotonic} belief revision.  
\subsection{Answer Set Programming(ASP)}
Answer Set Programming \cite{GEBSER_KAMINSKI_KAUFMANN_SCHAUB_2019} is a declarative programming paradigm that provides a natural and computationally efficient implementation of non-monotonic reasoning. ASP consists of rules of the form presented in equation \ref{eq:asp_rule}.

\begin{equation}
h \leftarrow b_1 \wedge b_2 \wedge \ldots 
\wedge b_m \wedge \neg c_1 \wedge \ldots 
\wedge \neg c_n
\label{eq:asp_rule}
\end{equation}
where $h$ is the head atom or conclusion, $b_1, \ldots, b_m$ are positive body literals and $\neg c_1, \ldots, \neg c_n$ encode negation as failure meaning the rule fires only if $c_1, \ldots, c_n$ cannot be derived from the current knowledge base. Clingo is the state-of-the-art ASP solver used in this work, supporting efficient grounding and solving of large-scale 
ASP programs.
\subsection{Machine Learning vs Non-Monotonic Reasoning}

Machine learning and non-monotonic reasoning systems presents fundamentally different paradigms of decision making. A machine learning classifier learns a fixed decision function from labeled training data as shown in equation \ref{eq:ml_function}. Once trained, this function is frozen and the classifier applies the same mapping to every new instance regardless of any additional contextual evidence available at the inference time. Incorporating new domain knowledge needs full model retraining incurring a computational cost. 
\begin{equation}
f: \mathcal{X} \rightarrow \mathcal{Y}, 
\quad f(\mathbf{x}) = \hat{y}, \quad 
\hat{y} \in \{\text{phishing}, 
\text{legitimate}\}
\label{eq:ml_function}
\end{equation}
Furthermore, ML classifiers provides no formal mechanism for retracting a prediction, for example, a false positive once generated cannot be revised without retraining. \\
Non-monotonic reasoning, by contrast, operates over an explicit knowledge base of facts and defeasible rules. When new contextual evidences becomes available, it is encoded as a new fact and reasoning engines re-evaluates the knowledge base to derive revised conclusions. This process operates in $\mathcal{O}(n)$ time and requires no modification to the underlying knowledge base structure. New knowledge is incorporated simply by adding new defeasible rules.
Table \ref{tab:ml_vs_nmr} summerizes the key differences between the two systems. This contrast motivates the notion of \textit{defeasible machine learning}, which bridges the gap between pattern recognition capability of statistical ML and the formal belief revision capability of NMR.

\subsection{Defeasible Machine Learning}

The term \textit{Defeasible machine learning is introduced by \cite{madzime2026defeasible}}. Our work extends this notion to statistical machine learning, treating the predictions of trained classifiers as defeasible initial beliefs subject to formal revision by an ASP-based non-monotonic reasoning layer. 

Conventional ML classifiers produce non-defeasible 
predictions that cannot be retracted 
without retraining. \textit{Defeasible machine learning} reframe this paradigm by treating classifier predictions not as fixed outputs but as \textit{defeasible initial beliefs} $\mathcal{B}_0$, that are subject to formal revision when contextual evidence supports a different conclusions. Formally, a defeasible machine learning system comprises two components: a statistical learning component that generates initial beliefs from input features and a non-monotonic reasoning component that formally revises these beliefs using domain-derived defeasible rules as presented in equation \ref{eq:dml}. 

\begin{equation}
\mathcal{B}_f = 
\text{NMR}(\mathcal{B}_0, 
\mathcal{E}, \mathcal{R})
\label{eq:dml}
\end{equation}

where $\mathcal{B}_0$ is the initial belief generated by the ML classifier, $\mathcal{E}$ is the set of contextual 
evidence facts, and $\mathcal{R}$ is the set of defeasible rules encoding domain knowledge. The final belief 
$\mathcal{B}_f$ is the result of applying the NMR reasoning layer to $\mathcal{B}_0$ under the evidence $\mathcal{E}$ and rules $\mathcal{R}$. This paradigm is applicable to any classification task where contextual evidence is available at inference time, false positives carry significant operational costs and new domain knowledge must be incorporated without retraining.

\begin{table}[h]
\centering
\caption{Comparison of Machine Learning 
and Non-Monotonic Reasoning}
\label{tab:ml_vs_nmr}
\small
\begin{tabular}{|p{3.5cm}|p{3.8cm}|p{3.5cm}|}
\hline
\textbf{Property} & 
\textbf{ML Classifier} & 
\textbf{NMR (ASP)} \\
\hline
Learning & 
Data-driven & 
Rule-driven \\
\hline
Decision boundary & 
Fixed after training & 
Dynamic via rules \\
\hline
Belief revision & 
Not supported & 
Formally supported \\
\hline
New evidence & 
Requires retraining & 
Rule addition only \\
\hline
Update complexity & 
$\mathcal{O}(T \cdot n \cdot f 
\cdot \log n)$ & 
$\mathcal{O}(n)$ \\
\hline
Defeasibility & 
No & 
Yes \\
\hline
Interpretability & 
Limited & 
Full --- rule-based \\
\hline
False positive control & 
Implicit via threshold & 
Explicit via rules \\
\hline
\end{tabular}
\end{table}

\subsection{Role of NMR for Phishing Detection}
The characteristics of phishing detection make it particularly well suited to nmr based post-hoc reasoning. First, phishing classifier trained on lexical features, are known to generate false positives that is legitimate URLs that share structural characteristics with phishing pages are incorrectly flagged as malicious~\cite{ban2023breaking}. These false positives cannot be corrected or altered without retraining the classifier. Second, new phishing strategies emerge continuously, requiring rapid incorporation of new detection signals without the overhead of full model retraining. Third, phishing detection systems operates in adversarial environments where phishing pages deliberately mimic legitimate structural characteristics which demands contextual , evidence driven revision of input classifier predictions rather than purely statistical classification. 

The proposed framework, addresses all three requirements by positioning a post-hoc ASP-based NMR layer after the machine learning classification phase. The NMR layer accepts initial classifier predictions as default beliefs $\mathcal{B}_0$ and formally revises them using EDA-derived defeasible rules when contextual evidence supports a different conclusion, presented in \ref{eq:belief_revision}, where $\mathcal{B}_0$ is the initial 
classifier belief, $M$ = 
\textit{meta\_info\_present}, 
$Q_1$ = \textit{long\_url} 
(\textit{length\_url} $> 63$), 
$Q_2$ = \textit{no\_www} 
(\textit{nb\_www} $= 0$), and 
$Q_3$ = \textit{domain\_title} 
(\textit{domain\_in\_title} $= 1$)

\begin{equation}
\mathcal{B}_f =
\begin{cases}
\text{legitimate}, & \text{if } 
\mathcal{B}_0 = \text{phishing} 
\wedge M = \text{yes} \wedge 
\neg Q_1 \wedge \neg Q_2 \wedge 
\neg Q_3 \\
\mathcal{B}_0, & \text{otherwise}
\end{cases}
\label{eq:belief_revision}
\end{equation}

This enables selective and evidence driven belief correction in $\mathcal{O}(n)$ time without modifying or retraining the underlying ML classifiers.

\section{Dataset information and feature engineering}\label{datasetinfo}
We experimented on the dataset available at \cite{hannousse2021web} which contains total 11,430 URLs. 50\% of 11,430 that is 5,715 URLs are of phishing type and remaining 50\% (5715) are of legitimate type. The dataset has been widely adopted as a standard benchmark for phishing detection research works with multiple state-of-art studies \cite{adane2023single, shafin2025explainable}. 
\subsection{Feature Set}
The dataset contains 87 features extracted from each URL, covering four broad categories as mentioned below.
\begin{itemize}
\item URL Structure Features\\
Lexical and syntactical characteristics of the raw URL string, including URL length, number of dots, number of hyphens in URL, presence of special character in URL, length of subdomain path etc.
\item Web Content Features\\
Characteristics extracted from the webpage content, including the number of hyperlinks, presence of forms, iframe usage, domain embedded in title page etc. 
\item Domain Based Features\\
Characteristics derived from the domain registration information including domain age, domain registration length etc.
\item External Service Features\\
Features derived from external services like Google index status, page rank. 

\end{itemize}
Along with these 87 features, this work introduces a novel binary feature named \textit{meta\_info\_present} which captures the presence or absence of meta tag information in the webpage associated with each URL. This feature was constructed by independently parsing each URL's webpage and recording the binary availability of meta fields such as as \texttt{description}, \texttt{keywords}, and \texttt{author}. The feature is encoded as 
\textit{yes} if meta information is 
present and \textit{no} otherwise.

\section{Exploratory Data Analysis of Contextual Features}
\label{eda}
This section presents the exploratory data analysis(EDA)  performed to extract the features used in the NMR reasoning layer. We analyse four features \textit{meta\_info\_present}, \textit{length\_url}, \textit{nb\_www}, and \textit{domain\_in\_title}. The EDA finding motivates the selection of the four rules of the proposed NMR framework.

\subsection{meta\_info\_present}
The binary feature \textit{meta\_info\_present} captures the presence or absence of meta tag 
information in the webpage associated with each URL. Meta tags such as \texttt{description}, 
\texttt{keywords}, and \texttt{author} are standard elements of legitimate web development practice, 
commonly used to support search engine indexing and semantic web standards. Phishing pages, which are 
typically hastily constructed to mimic legitimate pages, often omit these structural elements. This 
feature was constructed by independently parsing each URL's webpage and recording the binary 
availability of meta fields. In the training set, 50.72\% of legitimate URLs contained meta 
information compared to only 10.0\% of phishing URLs. This is a separation of 40.72 percentage 
points --- confirming its discriminative value as the primary contextual signal for belief revision. A consistent pattern is observed in the test set, where 49.69\% of legitimate URLS contain meta information compared to only 9.80\% of phishing URLs. The test set values are reported for verification only and they were not used to build the rules. Table \ref{tab:meta_distribution} shows the detail distribution of meta tags. This feature motivates \textbf{Rule 1} of the NMR framework: when a classifier predicts phishing and meta information is present, the prediction is revised to legitimate, subject to override by Rules 2--4.

\begin{table}[h]
\centering
\caption{Distribution of \textit{meta\_info\_present} across the 
training and test sets. Training-set values are used for rule 
derivation; test-set values are shown for verification only.}
\label{tab:meta_distribution}
\small
\begin{tabular}{|l|l|c|c|c|}
\hline
\textbf{Split} & 
\textbf{URL Type} & 
\textbf{Meta Present} & 
\textbf{Meta Absent} & 
\textbf{Meta (\%)} \\
\hline
\multirow{2}{*}{\textbf{Train}} & 
Legitimate (4,572) & 2,319 & 2,253 & 50.72 \\
\cline{2-5}
& Phishing (4,570) & 457 & 4,113 & 10.00 \\
\hline
\multicolumn{2}{|c|}{\textbf{Total (9,142)}} & 
\textbf{2,776} & \textbf{6,366} & 
\textbf{30.37} \\
\hline
\multirow{2}{*}{\textbf{Test}} & 
Legitimate (1,143) & 568 & 575 & 49.69 \\
\cline{2-5}
& Phishing (1,143) & 112 & 1,031 & 9.80 \\
\hline
\multicolumn{2}{|c|}{\textbf{Total (2,286)}} & 
\textbf{680} & \textbf{1,606} & 
\textbf{29.75} \\
\hline
\end{tabular}
\end{table}

\subsection{length\_url}

\begin{table*}[h]
\centering
\caption{Descriptive Statistics and Percentile 
Comparison of length\_url (Training Set) --- 
Phishing (meta=yes, $n$=457) vs 
Legitimate (meta=yes, $n$=2,319)}
\label{tab:url_length_stats}
\begin{tabular}{|l|c|c||l|c|c|}
\hline
\multicolumn{3}{|c||}{\textbf{Descriptive Statistics}} & 
\multicolumn{3}{c|}{\textbf{Percentile Comparison}} \\
\hline
\textbf{Metric} & 
\textbf{Phishing} & 
\textbf{Legitimate} & 
\textbf{Percentile} & 
\textbf{Phishing} & 
\textbf{Legitimate} \\
\hline
Count & 457 & 2,319 & P10 & 27.0 & 23.0 \\
Mean & 63.22 & 45.47 & P25 & 34.0 & 27.0 \\
Std Dev & 49.59 & 23.09 & P50 & 47.0 & 38.0 \\
Min & 15.0 & 13.0 & P75 & 74.0 & 59.0 \\
Max & 415.0 & 161.0 & P90 & 113.4 & 77.0 \\
Median & 47.0 & 38.0 & P95 & 166.2 & 90.0 \\
\hline
\end{tabular}
\end{table*}

\begin{figure*}[!t]
\centering
\includegraphics[height=8cm, width=13cm]{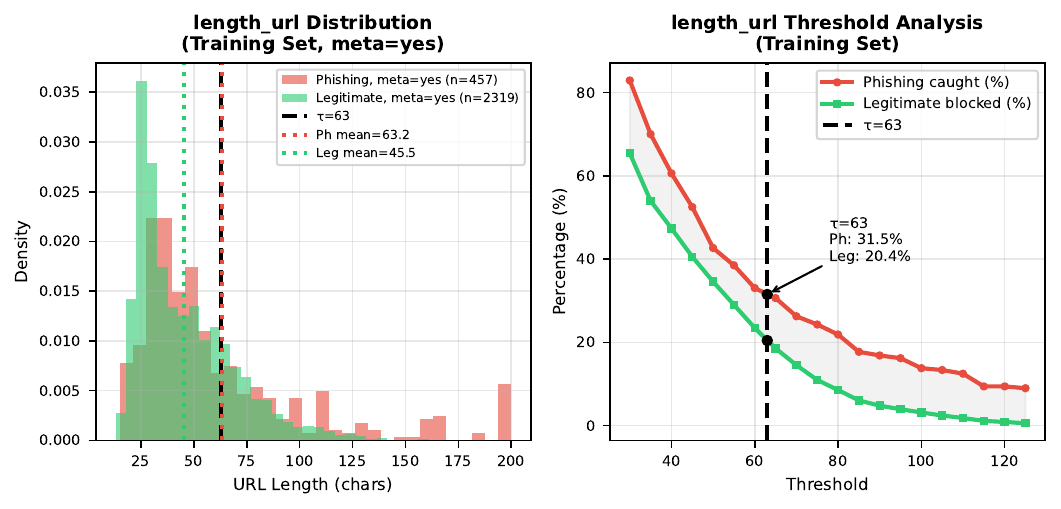}
\caption{Distribution of length\_url in the training set for instances with meta=yes (phishing $n = 457$, legitimate $n = 2{,}319$). The left panel shows the density histogram for both classes. The dashed black line marks the threshold $\tau = 63$. The dotted lines mark the class means. URL lengths above 200 characters are shown in the last bin. The phishing mean (63.2) lies very close to $\tau = 63$, so the two lines overlap. The right panel shows the percentage of URLs in each class that exceed each threshold. At $\tau = 63$, 31.5\% of phishing URLs and 20.4\% of legitimate URLs exceed the threshold.}
\label{fig:url_length}
\end{figure*}

URL length serves as a structural indicator of phishing behaviour. 
An examination over the URL length over the training set shows a distributional difference between phishing URLs with meta information present($n$ = 457) and legitimate with meta information present  ($n$ = 2319). On average, phishing URLs are 63.22 characters long with median of 47 while legitimate URLs have an average of 45.47 characters ($\sigma$=23.09 and phishing $\sigma$ = 49.59) with median of 38.0. The percentile analysis further confirms that the distribution gap widens at higher percentile. Phishing URLs are substantially longer  at P75 (74.0 vs 59.0), P90 (113.4 vs 77.0) P95 (166.2 vs 90.0). The descriptive statistics and percentile analysis both are presented in 
table \ref{tab:url_length_stats}. Figure \ref{fig:url_length} depicts two complementary interpretation of URL length feature. The left panel shows the density distribution. Legitimate URLs cluster densely at shorter lengths, while phishing URLs have a wider spread and a longer right tail. No legitimate URLs in the training set exceeds 161 characters, while the longest phishing URL has 415 characters. The threshold efficacy in the right panel shows the trade-off between phishing URLs caught and legitimate URLs blocked at various thresholds. At the chosen threshold $\tau$ = 63, which is roughly equal to the mean of phishing cases, 31.5\% of phishing URLs exceed the threshold while only 20.4\% of legitimate URLs exceed it, hence achieving a discrimination ratio of ~1.54 approximately. This motivates the construction of rule 2 that is when a URL is predicted as phishing has meta information present but exhibits excessive length ($>$ 63 characters), the revision to legitimate is defeated.

\subsection{nb\_www}
The feature \textit{nb\_www} records the number of occurrences of the sub-string \texttt{www} within the full URL string. Legitimate websites commonly adapt the general convention like \texttt{www.domain.com} which results the count of \texttt{www}=1. But, phishing websites frequently deviate from this convention by omitting \textit{www} prefix. Phishers often rely on IP addresses, obfuscated subdomains to conceal the fraudulent nature of a URL. This feature is analyzed only for instances with meta information present and length\_url $\leq 63$. These are the instances that reach Rule~3, because Rule~2 already defeats revision for longer URLs.

\begin{table}[h]
\centering
\caption{Distribution of \textit{nb\_www} (Training Set) --- 
Phishing vs Legitimate 
(meta=yes, length\_url $\leq$ 63)}
\label{tab:nb_www_stats}
\begin{tabular}{|l|c|c|}
\hline
\textbf{nb\_www Value} & 
\textbf{Phishing ($n$=313)} & 
\textbf{Legitimate ($n$=1,845)} \\
\hline
0 (no www prefix) & 243 (77.6\%) & 459 (24.9\%) \\
1 (has www prefix) & 69 (22.0\%) & 1,386 (75.1\%) \\
$\geq$ 2 & 1 (0.3\%) & 0 (0.0\%) \\
\hline
\textbf{Separation} & 
\multicolumn{2}{c|}{\textbf{52.8\%}} \\
\hline
\end{tabular}
\end{table}

Table \ref{tab:nb_www_stats} presents the distribution of \textit{nb\_www} in the training set. Among phishing URLs, 77.6\% omit the \texttt{www} prefix compared to 24.9\% of legitimate URLs. Conversely, 75.1\% of legitimate URLs contain the \texttt{www} prefix, against 22.0\% of phishing URLs. The separation of 52.8 percentage points is the largest among the four features. A chi-square test compares nb\_www = 0 against nb\_www $\geq 1$. The difference is statistically significant ($\chi^2 = 336.96$, df = 1, $p = 2.92 \times 10^{-75}$). Phishing URLs in this subset are about 10.5 times more likely to lack the \texttt{www} prefix. The odds ratio is 10.48. This finding motivates rule 3. When a URL is predicted as phishing and has meta information present, but has no \texttt{www} prefix, the revision to legitimate is defeated. 

\subsection{domain\_in\_title}
The feature \textit{domain\_in\_title} indicates whether the domain name of the URL is embedded within the webpage title. Legitimate websites commonly include their brand or domain name in the page title as part of standard web development practice. Table \ref{tab:domain_in_title_stats} presents the distribution of \textit{domain\_in\_title} for both classes of URLs in the training set. It covers the instances that reach rule 4, that is meta information present, length\_url $\leq 63$ and a \texttt{www} prefix. This subset has 69 phishing and 1386 legitimate URLs. In this subset, the separation is of 25.8 percentage points between phishing(82.6\%) and legitimate(56.8\%) entries that contains domain\_in\_title. A chi-square test confirms that the difference is statistically significant($\chi^2 = 16.95$, df = 1, $p = 3.83 \times 10^{-5}$). The odds ratio is 3.62. Phishing URLs in this subset are about 3.6 times more likely to have the domain in the title. This pattern is counter intuitive, because a domain in the title is usually seen as a sign of  legitimacy. It is consistent with the hypothesis that phishing pages may deliberately reproduce legitimate page-title characteristics to appear credible. Dataset alone cannot confirm attacker intent. This result motivates rule 4. When a URL is predicted as phishing and has meta information present, a short URL and a \texttt{www} prefix, but the domain appears in the title, the revision to legitimate is defeated. 

\begin{table}[h]
\centering
\caption{Distribution of \textit{domain\_in\_title} (Training Set) --- 
Phishing vs Legitimate 
(meta=yes, length\_url $\leq$ 63, nb\_www = 1)}
\label{tab:domain_in_title_stats}
\begin{tabular}{|l|c|c|}
\hline
\textbf{domain\_in\_title} & 
\textbf{Phishing ($n$=69)} & 
\textbf{Legitimate ($n$=1,386)} \\
\hline
0 (not in title) & 12 (17.4\%) & 599 (43.2\%) \\
1 (in title) & 57 (82.6\%) & 787 (56.8\%) \\
\hline
\textbf{Separation} & 
\multicolumn{2}{c|}{\textbf{25.8\%}} \\
\hline
\end{tabular}
\end{table}

\subsection{Summary of EDA Findings}
\begin{table*}[h]
\centering
\caption{EDA Summary --- Contextual Feature 
Separation Analysis Across All 4 Features
(Training Set).Each feature is measured on the subset that reaches its rule: meta\_info\_present on all training instances, length\_url on meta=yes, nb\_www on meta=yes with length\_url $\leq 63$, and domain\_in\_title on meta=yes with length\_url $\leq 63$ and nb\_www = 1.}
\label{tab:eda_summary}
\begin{tabular}{|p{3.7cm}|p{1.8cm}|p{2cm}|p{2.01cm}|p{3.3cm}|}
\hline
\textbf{Feature} & 
\textbf{Phishing (\%)} & 
\textbf{Legitimate (\%)} & 
\textbf{Separation (\%)} & 
\textbf{NMR Usage} \\
\hline
meta\_info\_present & 
10.00 & 50.72 & 40.72 & 
Rule 1 (revision trigger) \\
\hline
length\_url ($>$63) & 
31.5 & 20.4 & 11.1 & 
Rule 2 (defeater) \\
\hline
nb\_www (=0) & 
77.6 & 24.9 & 52.8 & 
Rule 3 (defeater) \\
\hline
domain\_in\_title (=1) & 
82.6 & 56.8 & 25.8 & 
Rule 4 (defeater) \\
\hline
\end{tabular}
\end{table*}

Table~\ref{tab:eda_summary} summarizes the separation analysis across all four features.
The separations range from 11.1 percentage points for \textit{length\_url} to 52.8 percentage points for \textit{nb\_www}.\textit{nb\_www} is the strongest binary signal. All four differences are statistically significant at $\alpha = 0.05$. No single feature separates the two classes on its own. Phishing and legitimate URLs overlap  on every feature. \textit{meta\_info\_present} is the revision trigger. Legitimate URLs are about 9.26 times more likely to have meta information. The other three features act as defeaters. Each one is applied to the subset left by the rules before it. This finding motivates, the need of multi-rule NMR framework that combines multiple weak signals as hierarchical defeasible rules within the ASP framework. On the test set, this combination revises five classifier predictions, and all five revisions are correct. 

\subsection{Statistical Significance of Contextual Features}
To address RQ1, the distributional differences between phishing and legitimate URLs identified through EDA were validated using statistical tests. For binary features like \textit{meta\_info\_present}, \textit{nb\_www} and \textit{domain\_in\_title}, Pearson's Chi-square test was performed with odds ratios as effect size measures (equation \ref{eq:chisquare}). A significance level of $\alpha = 0.05$ 
was adopted as the threshold for rejecting the null hypothesis, consistent with standard practice in statistical hypothesis testing. For each features, the null hypothesis $H_0$ states that the feature distribution is independent of the class type(phishing or legitimate) i.e the observed distributional differences arise by chance. The alternative hypothesis $H_1$ states that a meaningful association exists between the feature distribution and class type. Results of the statistical tests justifies its use as a defeasible revision signal in the NMR reasoning layer. For binary features --- 
\textit{meta\_info\_present}, \textit{nb\_www}, and \textit{domain\_in\_title} --- 
Pearson's Chi-square test of independence was applied. 

\begin{equation}
\chi^2 = \sum \frac{(O_{ij} - E_{ij})^2}
{E_{ij}}
\label{eq:chisquare}
\end{equation}

where $O_{ij}$ denotes the observed 
frequency and $E_{ij}$ denotes the 
expected frequency under $H_0$, 
computed as:

\begin{equation}
E_{ij} = \frac{(\text{row total}_i) 
\times (\text{column total}_j)}
{N}
\label{eq:expected}
\end{equation}  
where $N$ is the total sample size. Odds ratios (OR) are reported as effect size measures to assess practical significance, as p-values 
alone can be inflated by large sample 
sizes.\\
For the continuous feature \textit{length\_url}, the non-parametric Mann-Whitney U test was applied, with $r$ as the effect size measure. 
\begin{equation}
U = n_1 n_2 + \frac{n_1(n_1+1)}{2} 
- R_1
\label{eq:mannwhitney}
\end{equation}

where $n_1$ and $n_2$ are the 
sample sizes of the phishing and 
legitimate groups respectively, 
and $R_1$ is the sum of ranks 
for the phishing group. The 
effect size $r$ is computed from the standardised test statistic as shown in equation \ref{eq:effectsize}.

\begin{equation}
r = \frac{Z}{\sqrt{n_1 + n_2}}
\label{eq:effectsize}
\end{equation}

Table~\ref{tab:stat_tests} presents the statistical test results for all four features. All four features reject $H_0$ at 
$\alpha$ = 0.05, confirming statistically significant distributional differences between phishing and legitimate URLs. 
\begin{itemize}
    \item \textit{meta\_info\_present} Among the training instances, 457 out of 4570 phishing URLs (10.00\%) contained meta information where 2319 out of 4,572 legitimate 
URLs (50.72\%) contains meta fields (as shown in table \ref{tab:meta_distribution})  which states a separation of 40.72 percentage points. The Chi-square test confirms this difference is highly significant ($\chi^2$ = 1790.46, $p < 0.001$), with legitimate URLs 9.26 times 
more likely to contain meta tags 
than phishing URLs (OR = 9.26).
\item \textit{length\_url} 
Among phishing 
URLs with meta information present 
($n$ = 457), the mean URL length 
was 63.22 characters 
($\sigma$ = 49.59, median = 47.00), 
compared to 45.47 characters 
($\sigma$ = 23.09, median = 38.00) 
for legitimate URLs ($n$ = 2,319) 
--- a mean difference of 17.75 
characters(as presented in table \ref{tab:url_length_stats}). The Mann-Whitney U test 
confirms this difference is 
statistically significant 
($U$ = 650,657, 
$p = 1.23 \times 10^{-14}$), 
with a small effect size 
($r$ = 0.146), consistent with 
the known distributional overlap 
between phishing and legitimate 
URL lengths. This reflects its 
role as a weak standalone 
discriminator that nonetheless 
provides a meaningful defeater 
signal within the NMR framework.
\item \textit{nb\_www}
Among the subset of instances where meta information is present and URL length does not exceed $\tau$ = 63 characters, 243 out of 313 phishing URLs (77.6\%) lacked the \texttt{www} 
prefix (\textit{nb\_www} = 0) 
compared to only 459 out of 1845 legitimate URLs(24.9\%) indicates a clear separation of 52.8 percentage points. The Chi-square test 
confirms this is highly significant 
($\chi^2$ = 336.96, 
$p = 2.92 \times 10^{-75}$, 
OR = 10.48), indicating phishing 
URLs are 10.48 times more likely 
to lack the \texttt{www} prefix.

\item \textit{domain\_in\_title} Among 
the subset where meta information 
is present, URL length does not 
exceed $\tau$ = 63 characters, 
and \texttt{www} prefix is present, 
57 out of 69 phishing URLs (82.6\%) 
embedded the domain in the page 
title (\textit{domain\_in\_title} 
= 1), compared to 787 out of 1,386 
legitimate URLs (56.8\%) --- a 
separation of 25.8 percentage 
points. The Chi-square test 
confirms this association is 
statistically significant 
($\chi^2$ = 16.95, 
$p = 3.83 \times 10^{-5}$, 
OR = 3.62), indicating phishing 
pages in this subset are 3.62 
times more likely to embed the 
domain in the page title. These 
results collectively confirm that 
all four EDA-derived contextual 
signals are statistically grounded 
and directly address \textbf{RQ1}.
\end{itemize}
The critical value of the $\chi^2$ test at $\alpha$ = 0.05 
with $df$ = 1, obtained from the standard Chi-square distribution table is $\chi^2_{critical}$ = 3.841 as presented in table \ref{tab:stat_tests}. All computed $\chi^2$ statistics substantially exceed this critical value, confirming the rejection of $H_0$ in all cases. 
\begin{table}[h]
\centering
\caption{Statistical Significance of 
Contextual Features --- Training Set
($\alpha$ = 0.05, df=1)}
\label{tab:stat_tests}
\small
\begin{tabular}{|p{3.1cm}|p{1.3cm}|p{1.4cm}|p{1.3cm}|p{1.4cm}|p{1.6cm}|c|}
\hline
\textbf{Feature} & 
\textbf{Test} & 
\textbf{Statistic} & 
\textbf{Critical} &
\textbf{p-value} & 
\textbf{Effect} &
\textbf{Decision} \\
\hline
\textit{meta\_info\_present} & 
$\chi^2$ & 
1790.46 & 
3.841 &
$<$0.001 & 
OR=9.26 &
Reject $H_0$ \\
\hline
\textit{length\_url} & 
Mann-Whitney & 
650657 & 
--- &
$1.23\times10^{-14}$ & 
$r$=0.146 &
Reject $H_0$ \\
\hline
\textit{nb\_www} & 
$\chi^2$ & 
336.96 & 
3.841 &
$2.92\times10^{-75}$ & 
OR=10.48 &
Reject $H_0$ \\
\hline
\textit{domain\_in\_title} & 
$\chi^2$ & 
16.95 & 
3.841 &
$3.83\times10^{-05}$ & 
OR=3.62 &
Reject $H_0$ \\
\hline
\end{tabular}
\end{table}

\section{Proposed Framework: PHISHREV+}\label{sec:framework}
\begin{figure}[htbp]
  \centering
  \includegraphics[width=\linewidth, height=9cm]{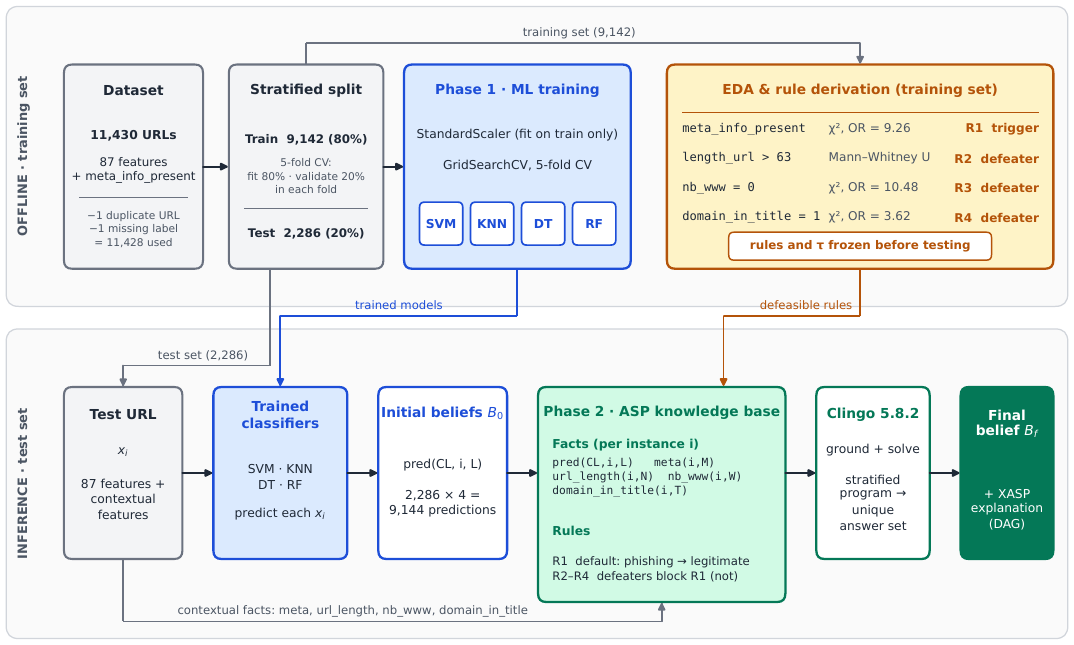}
  \caption{Overview of the PHISHREV+ framework. In the first stage, the training set is used to train four ML classifiers and to derive the four defeasible rules and the threshold $\tau = 63$ through EDA. The rules and $\tau = 63$ are frozen before testing. In the inference stage, each test URL is classified by the trained models. These predictions become initial beliefs $B_0$. These beliefs and the contextual features of the URL are encoded as ASP facts. Clingo computes the unique answer set, which gives the final belief $B_f$.  }
  \label{fig:workflow}
\end{figure}

The proposed framework operates in two sequential processing stages- an initial machine learning classification stage that generates preliminary predictions referred to as initial beliefs and a subsequent NMR belief revision phase that selectively revises these predictions using EDA-derived evidence encoded as defeasible ASP rules. Figure \ref{fig:workflow}, illustrates the complete architecture of the proposed framework.  

\subsection{Phase 1: Machine Learning Classification}
Four machine learning classifiers namely Support Vector Machine (SVM), K-Nearest Neighbour (KNN), Decision Tree (DT), and Random Forest (RF) were used.
\subsubsection{Data Preparation}
The dataset contains 11,430 rows \cite{hannousse2021web}. One URL appears twice which leaves 11,429 unique URLs. These are split into
  training set of 9,143 instances (80\%) and test set of 2,286 samples(20\%) using stratified splitting. One training instance had a missing class label and was removed. The final data therefore contains 11,428 instances : 9,142 for training and 2,286 for testing. All features were normalized using Standard Scaler technique as shown in algorithm \ref{algo:phase1}. 
  
\begin{algorithm}[h]
\caption{PHISHREV+ Data Preparation}
\label{algo:phase1}
\begin{algorithmic}[1]
\Require Dataset $\mathcal{D}$ with 
         87 benchmark features, 
         1 novel feature 
         \textit{meta\_info\_present} 
         and a class label 
         $y \in \{\text{phishing}, \text{legitimate}\}$ 
         for each URL,
         Classifiers $\mathcal{C}$ = 
         \{SVM, KNN, DT, RF\},
         EDA-derived threshold 
         $\tau = 63$,
         ASP solver Clingo
\Ensure Final belief $\mathcal{B}_f$ 
        for each test instance
\Statex
\State Partition $\mathcal{D}$ into 
       $\mathcal{D}_{train}$ (80\%, 
       9,142 instances after cleaning) and 
       $\mathcal{D}_{test}$ (20\%, 
       2,286 instances) using 
       stratified splitting
\State Apply Standard Scaler 
       normalisation to 
       $\mathcal{D}_{train}$ and 
       $\mathcal{D}_{test}$, with 
       parameters fitted on 
       $\mathcal{D}_{train}$ only 
       to prevent data leakage
\end{algorithmic}
\end{algorithm}

\subsubsection{Classifier Training and Hyperparameter Optimisation}
Hyperparameter optimisation is performed using GridSearchCV with 5-fold cross validation on the training set. The search space and best hyperparameters are placed in table \ref{tab:hyperparams}.

\begin{table}[h]
\centering
\caption{Best Hyperparameter Configuration for Each Classifier}
\label{tab:hyperparams}
\small
\begin{tabular}{|p{1.7cm}|p{5cm}|p{3.8cm}|}
\hline
\textbf{Classifier} & 
\textbf{Search Space} & 
\textbf{Best Configuration} \\
\hline
SVM & 
C $\in$ \{0.1, 1, 10\}, 
kernel $\in$ \{linear, rbf\}, 
$\gamma$ $\in$ \{scale, auto\} & 
C=10, kernel=rbf, 
$\gamma$=scale \\
\hline
KNN & 
k $\in$ \{3,5,7,9\}, 
weights $\in$ \{uniform, distance\}, 
metric $\in$ \{euclidean, manhattan\} & 
k=7, weights=distance, 
metric=manhattan \\
\hline
DT & 
criterion $\in$ \{gini, entropy\}, 
max\_depth $\in$ \{3,5,10,None\}, 
min\_samples\_split $\in$ \{2,5,10\} & 
criterion=gini, 
max\_depth=10, 
min\_samples\_split=2 \\
\hline
RF & 
n\_estimators $\in$ \{100,200\}, 
max\_depth $\in$ \{5,10,None\}, 
criterion $\in$ \{gini, entropy\} & 
n\_estimators=200, 
max\_depth=None, 
criterion=gini \\
\hline
\end{tabular}
\end{table}

\subsubsection{Initial Belief Generation}
Each trained classifier is evaluated on the test set to generate initial predictions. These predictions are treated as initial beliefs $\mathcal{B}_0$ in the NMR reasoning layer as shown in equation \ref{eq:initial_belief}.

\begin{equation}
\mathcal{B}_0(CL, i) = CL(\mathbf{x}_i) 
\in \{\text{phishing, legitimate}\}
\label{eq:initial_belief}
\end{equation}
where, $CL \in$ \{SVM, KNN, DT, RF\} denotes the classifier and $\mathbf{x}_i$ denotes the $i$-th test instance. These initial beliefs represent the closed world predictions of the ML classifications prior to any revision. Algorithm \ref{algo:phase2}, presents the steps.

\begin{algorithm}[h]
\caption{PHISHREV+: ML Classification and Initial Belief Generation}
\label{algo:phase2}
\begin{algorithmic}[1]
\setcounter{ALG@line}{2}
\State Train classifiers 
       $\mathcal{C}$ = 
       \{SVM, KNN, DT, RF\} on 
       $\mathcal{D}_{train}$ using 
       5-fold cross-validation and 
       GridSearchCV for hyperparameter 
       optimisation
\For{each instance 
     $\mathbf{x}_i \in 
     \mathcal{D}_{test}$}
    \For{each classifier 
         $CL \in \mathcal{C}$}
        \State Generate initial belief:
               $\mathcal{B}_0(CL, i) 
               \leftarrow 
               CL(\mathbf{x}_i) \in 
               \{\text{phishing}, 
               \text{legitimate}\}$
    \EndFor
\EndFor
\end{algorithmic}
\end{algorithm}

\subsection{Phase 2: ASP-Based Non-Monotonic Reasoning Layer}

In this phase, the knowledge base is created and used for the belief revision. 
\subsubsection{Knowledge Base Construction}
The NMR reasoning layer operates over an ASP knowledge base comprising two components: a. ground facts encoding the initial classifier predictions and contextual feature values and defeasible rules encoding the EDA-derived revision conditions. For each test instance, $x_i$, the following ASP facts are generated\\

\begin{itemize}
\item \texttt{pred(CL, i, Label).}  
encodes the initial belief of classifier 
$CL$ for instance $i$, where 
\texttt{Label} $\in$ 
\{phishing, legitimate\}

\item \texttt{meta(i, M).} encodes 
the meta tag availability for instance 
$i$, where \texttt{M} $\in$ \{yes, no\}

\item \texttt{url\_length(i, L).} 
encodes the URL length for instance 
$i$, where \texttt{L} is an integer

\item \texttt{nb\_www(i, W).} encodes the www prefix count for 
instance $i$, where \texttt{W} 
$\in$ \{0, 1\}

\item \texttt{domain\_in\_title(i, T).} 
--- encodes the domain-in-title signal 
for instance $i$, where \texttt{T} 
$\in$ \{0, 1\}

\end{itemize}
In algorithm \ref{algo:phase3}, the steps are outlined. A sample of the generated ASP facts for two representative instances is table \ref{tab:asp_facts}.

\begin{algorithm}[h]
\caption{PHISHREV+: ASP Knowledge Base Construction}
\label{algo:phase3}
\begin{algorithmic}[1]
\setcounter{ALG@line}{6}
\For{each instance 
     $\mathbf{x}_i \in 
     \mathcal{D}_{test}$}
    \State Generate ASP ground facts:
    \State \hspace{1em}
           \texttt{pred($CL$, $i$, 
           $\mathcal{B}_0(CL,i)$).}
           \Comment{initial belief}
    \State \hspace{1em}
           \texttt{meta($i$, 
           meta\_info\_present[$i$]).}
           \Comment{novel feature}
    \State \hspace{1em}
           \texttt{url\_length($i$, 
           length\_url[$i$]).}
           \Comment{Rule 2 signal}
    \State \hspace{1em}
           \texttt{nb\_www($i$, 
           nb\_www[$i$]).}
           \Comment{Rule 3 signal}
    \State \hspace{1em}
           \texttt{domain\_in\_title($i$, 
           domain\_in\_title[$i$]).}
           \Comment{Rule 4 signal}
\EndFor
\end{algorithmic}
\end{algorithm}

\begin{table}[!t]
\centering
\caption{ASP facts corresponding to two sample URLs}
\label{tab:asp_facts}
\begin{tabular}{|l|l|}
\hline
\textbf{Predicate} & \textbf{Fact} \\
\hline
Prediction & \texttt{pred(rf,464,phishing).} \\
Metadata & \texttt{meta(464,yes).} \\
URL Length & \texttt{url\_length(464,44).} \\
WWW Count & \texttt{nb\_www(464,1).} \\
Domain in Title & \texttt{domain\_in\_title(464,0).} \\
\hline
Prediction & \texttt{pred(svm,2048,phishing).} \\
Metadata & \texttt{meta(2048,yes).} \\
URL Length & \texttt{url\_length(2048,28).} \\
WWW Count & \texttt{nb\_www(2048,1).} \\
Domain in Title & \texttt{domain\_in\_title(2048,0).} \\
\hline
\end{tabular}
\end{table}

\subsubsection{Defeasible NMR Rules}
Four defeasible rules are encoded within the ASP knowledge base, derived directly from the EDA findings presented in section \ref{eda}. Rules interact hierarchically that is rule 1 initiates belief revision while rules 2 to 4 act as defeasible exceptions that defeat rule 1.\\

\textbf{Rule 1 (Base Revision Rule):} 
A classifier prediction of phishing is revised to legitimate when meta information is present and no adversarial defeater signal is detected. Equation \ref{eq:rule1} presents the rule. 
\begin{equation}
\begin{split}
\text{revised}(CL, i, \text{legitimate}) \leftarrow{}
    & \text{pred}(CL, i, \text{phishing}) \\
    & {}\wedge \text{meta}(i, \text{yes})
       \wedge \neg\,\text{long\_url}(i) \\
    & {}\wedge \neg\,\text{no\_www}(i)
       \wedge \neg\,\text{domain\_title}(i)
\end{split}
\label{eq:rule1}
\end{equation}

\textbf{Rule 2 (URL Length Defeater):} 
Derived from EDA finding in section \ref{eda}, that phishing URLs with meta information present have a mean length of 63.22 characters ($\sigma$ = 49.59) compared to 45.47 ($\sigma$ = 23.09) for legitimate URLs. At threshold($\tau$ = 63), 31.5\% of phishing instances exceed the threshold against 20.4\% of legitimate URLs. Equation \ref{eq:rule2}, states rule 2.   
\begin{equation}
\begin{split}
\text{revised}(CL, i, \text{phishing}) \leftarrow{}
    & \text{pred}(CL, i, \text{phishing}) \\
    & {}\wedge \text{meta}(i, \text{yes}) \\
    & {}\wedge \text{long\_url}(i)
\end{split}
\label{eq:rule2}
\end{equation}

\textbf{Rule 3 (www Defeater):} 
From section \ref{eda}, we found that 77.6\% of phishing instances with meta present and url of length $\leq$ 63 lack the \texttt{www} prefix compared to only 24.9\% of legitimate URLS that is the separation is of 52.8 percentage points. The rule is presented in equation \ref{eq:rule3}.
\begin{equation}
\begin{split}
\text{revised}(CL, i, \text{phishing}) \leftarrow{}
    & \text{pred}(CL, i, \text{phishing}) \\
    & {}\wedge \text{meta}(i, \text{yes})
       \wedge \neg\,\text{long\_url}(i) \\
    & {}\wedge \text{no\_www}(i)
\end{split}
\label{eq:rule3}
\end{equation}

\textbf{Rule 4 (Domain-Title Defeater):}
EDA \ref{eda} reveals among instances with meta information present, length\_url $\leq 63$ and a \texttt{www} prefix, 82.6\% of phishing URLs embed the domain in the page title, against 56.8\% of legitimate URLs. The separation is 25.8 percentage points. Rule 4 is presented in equation \ref{eq:rule4}.
\begin{equation}
\begin{split}
\text{revised}(CL, i, \text{phishing}) \leftarrow\ & \text{pred}(CL, i, \text{phishing}) \\
& \wedge\ \text{meta}(i, \text{yes}) \wedge \neg\, \text{long\_url}(i) \\
& \wedge\ \neg\, \text{no\_www}(i) \wedge \text{domain\_title}(i)
\end{split}
\label{eq:rule4}
\end{equation}

The final belief $\mathcal{B}_f$ for 
each instance is computed as presented in equation \ref{eq:final_belief}.
\begin{equation}
\mathcal{B}_f(CL, i) =
\begin{cases}
    \text{legitimate}, & \text{if } \text{revised}(CL, i, \text{legitimate}) \\
    \text{phishing},   & \text{if } \text{revised}(CL, i, \text{phishing}) \\
    \mathcal{B}_0(CL, i), & \text{otherwise}
\end{cases}
\label{eq:final_belief}
\end{equation}
The four rules interact hierarchically. Rules 2-4, act as defeasible exceptions that formally defeat rule 1 under conflicting contextual evidence - a behaviour encoded through the \texttt{not} operator in ASP, representing negation as failure. Rule 1 fires only when rule 2,3 and 4 fail to derive their conclusions, capturing the defeasible nature of the revision. The first two cases of equation \ref{eq:final_belief}, therefore cannot hold together, and each prediction has exactly one final belief. In algorithm \ref{algo:phase4}, steps are mentioned.

\begin{algorithm}[h]
\caption{PHISHREV+ --- NMR Belief Revision}
\label{algo:phase4}
\begin{algorithmic}[1]
\setcounter{ALG@line}{13}
\For{each instance 
     $\mathbf{x}_i \in 
     \mathcal{D}_{test}$}
    \For{each classifier 
         $CL \in \mathcal{C}$}
        \If{$\mathcal{B}_0(CL, i) = 
            \text{phishing} \wedge 
            \text{meta}(i) = \text{yes}$}
            \If{$\text{url\_length}(i) 
                > \tau$}
                \State \textbf{Rule 2:} 
                       $\mathcal{B}_f 
                       \leftarrow 
                       \text{phishing}$
                \Comment{long URL — 
                         Rule 2 defeats 
                         Rule 1}
            \ElsIf{$\text{nb\_www}(i) 
                   = 0$}
                \State \textbf{Rule 3:} 
                       $\mathcal{B}_f 
                       \leftarrow 
                       \text{phishing}$
                \Comment{no www — 
                         Rule 3 defeats 
                         Rule 1}
            \ElsIf{$\text{domain\_title}
                   (i) = 1$}
                \State \textbf{Rule 4:} 
                       $\mathcal{B}_f 
                       \leftarrow 
                       \text{phishing}$
                \Comment{domain in title —
                         Rule 4 defeats 
                         Rule 1}
            \Else
                \State \textbf{Rule 1:} 
                       $\mathcal{B}_f 
                       \leftarrow 
                       \text{legitimate}$
                \Comment{all conditions 
                         met — Rule 1 
                         fires}
            \EndIf
        \Else
            \State $\mathcal{B}_f(CL, i) 
                   \leftarrow 
                   \mathcal{B}_0(CL, i)$
            \Comment{no revision needed}
        \EndIf
    \EndFor
\EndFor
\end{algorithmic}
\end{algorithm}

\subsubsection{Computational Complexity}
Table \ref{tab:complexity}, presents the computational complexity of each phase of the proposed framework. A key practical advantage of the proposed work is that the total NMR overhead is $\mathcal{O}(n)$ that is linear in the number of test samples and independent of the number of features $f$ or number of trees $T$. Incorporating new domain knowledge requires only adding a new defeasible rule/logic to the ASP knowledge base, operating in $\mathcal{O}(n)$ time, compared to full model retraining.  

\begin{table}[h]
\centering
\caption{Computational Complexity of 
PHISHREV+ Framework}
\label{tab:complexity}
\small
\begin{tabular}{|p{3cm}|p{2cm}|p{2.5cm}|}
\hline
\textbf{Phase} & 
\textbf{Operation} & 
\textbf{Complexity} \\
\hline
ML Training & 
SVM & 
$\mathcal{O}(n^2 \cdot f)$ \\
\hline
ML Training & 
KNN & 
$\mathcal{O}(n \cdot f)$ \\
\hline
ML Training & 
DT & 
$\mathcal{O}(n \cdot f \cdot \log n)$ \\
\hline
ML Training & 
RF & 
$\mathcal{O}(T \cdot n \cdot f 
\cdot \log n)$ \\
\hline
NMR Fact Generation & 
All classifiers & 
$\mathcal{O}(n)$ \\
\hline
NMR Rule Evaluation & 
All classifiers & 
$\mathcal{O}(n)$ \\
\hline
\textbf{Total NMR Overhead} & 
\textbf{All classifiers} & 
$\mathcal{O}(n)$ \\
\hline
\textbf{Knowledge Update} & 
\textbf{New rule addition} & 
$\mathcal{O}(n)$ \\
\hline
\end{tabular}
\end{table}

Let $n$ denote the number of URL instances submitted for revision. The knowledge base comprises (i)~a bounded number
$k$ of ground facts per instance --- namely 
\texttt{pred(CL, ID, C)}, 
\texttt{meta(ID, V)}, 
\texttt{url\_length(ID, L)}, 
\texttt{nb\_www(ID, W)}, and 
\texttt{domain\_in\_title(ID, T)}  --- yielding $\mathcal{O}(n)$
ground facts in total, and (ii)~a fixed finite set $R$ of
defeasible revision rules whose cardinality is independent
of $n$.

\begin{proposition}
Re-evaluation of the proposed knowledge base under the
answer-set semantics runs in $\mathcal{O}(n)$ time.
\end{proposition}

\begin{proof}[Proof sketch]
The program is \emph{stratified \footnote{A logic program is \emph{stratified} when its rules can be ordered
into layers such that no conclusion depends negatively on
itself, ensuring a single, well-defined outcome; it is
\emph{non-disjunctive} when every rule derives exactly one
conclusion rather than a choice among alternatives. Together
these properties guarantee a unique answer set that can be
computed without search.}and non-disjunctive, with
revision rules applied independently per instance}. Three
properties follow.

\textbf{Grounding.} Every rule body ranges over a single
instance variable $ID$ together with locally bounded
variables ($CL$, $L$) drawn from fixed finite domains, and
no rule joins two distinct instance identifiers. Grounding
each of the $\lvert R\rvert$ rules against the facts therefore
yields at most $\mathcal{O}(n)$ ground instances, and the
absence of cross-instance joins precludes any $\mathcal{O}(n^{2})$
combinatorial blow-up. The ground program has size
$\mathcal{O}(\lvert R\rvert \cdot n) = \mathcal{O}(n)$.

\textbf{Uniqueness of the model.} Because negation-as-failure
is stratified (the dependency order 
\texttt{long\_url} $\rightarrow$ 
\texttt{revised} $\rightarrow$ 
\texttt{final}contains no negative cycle) and
the program contains no disjunction, the program admits a
\emph{single answer set}. Consequently no
search over competing extensions is required, and the
second-level complexity characteristic of general
non-monotonic formalisms does not arise.

\textbf{Solving.} The unique model is obtained by a bottom-up
evaluation of the strata in which each ground rule fires a
constant number of times. Since revision rules apply
independently per instance, the global answer set is the
disjoint union of $n$ per-instance computations, each of
$\mathcal{O}(1)$ cost. Solving is thus $\mathcal{O}(n)$.

Combining grounding and solving gives an overall cost of
$\mathcal{O}(n)$.
\end{proof}
Incorporating new domain knowledge amounts to appending a
rule to $R$, an $\mathcal{O}(1)$ edit that leaves the
re-evaluation cost at $\mathcal{O}(n)$ and requires no
retraining of the underlying classifiers. This tractability
is a property of the restricted rule fragment employed and
is not claimed for non-monotonic reasoning in general, whose
expressive fragments (disjunctive or unstratified programs)
are $\Sigma_{2}^{p}$-hard.

\section{Experimental Results}\label{result}
This section presents, the experimental evaluation of the proposed framework. The evaluation is structured in two parts: the first part consists of the machine learning classifiers before NMR-based belief revision and the second part contains the performance of after the NMR-based belief revision using the proposed defeasible framework.    

\subsection{Experimental Setup}
All experiments were conducted on a computer running Microsoft Windows 11 Education with an Intel Core i5-14400 processor and 16 GB RAM. The pipeline was implemented in Python 3.10 using scikit-learn~1.7.2 for machine learning and the Clingo 5.8.2 ASP solver, for non-monotonic reasoning.
\subsection{Phase 1: ML Classification Results — Initial Belief Generation}

Table \ref{tab:ml_results} presents the performance of the four machine learning classifiers on the test set prior to NMR-based belief revision. These predictions constitutes the initial beliefs $\mathcal{B}_0$ fed into the NMR reasoning layer.  

\begin{table*}[h]
\centering
\caption{Machine Learning Classification 
Results Before NMR-Based Belief Revision}
\label{tab:ml_results}
\small
\begin{tabular}{|l|c|c|c|c|c|c|}
\hline
\textbf{Classifier} & 
\textbf{Accuracy (\%)} & 
\textbf{Precision} & 
\textbf{Recall} & 
\textbf{F1-Score} & 
\textbf{FP} & 
\textbf{FN} \\
\hline
SVM & 96.33 & 0.9641 & 0.9624 & 
0.9632 & 41 & 43 \\
\hline
KNN & 95.45 & 0.9585 & 0.9501 & 
0.9543 & 47 & 57 \\
\hline
Decision Tree & 93.70 & 0.9448 & 
0.9283 & 0.9365 & 62 & 82 \\
\hline
Random Forest & 95.80 & 0.9548 & 
0.9615 & 0.9582 & 52 & 44 \\
\hline
\end{tabular}
\end{table*}

Among the four classifiers, SVM achieves the highest accuracy of 96.33\% followed by Random Forest at 95.80\%, KNN at 95.45\% and Decision Tree at 93.70\%. SVM achieved highest recall of 0.9624, indicating strong sensitivity to phishing instances. All four classifiers demonstrate competitive performance on the benchmark dataset, consistent with prior work\cite{hannousse2021towards, adane2023single, shafin2025explainable}. However, all classifiers generate false positives, ranging from 41(SVM) to 62(DT), representing legitimate URLs incorrectly flagged as phishing. These false positives constitute the primary target for NMR-based belief revision in Phase 2. \\
As depicted in table \ref{tab:meta_distribution}, approximately 50\% legitimate URLs contain meta tags compared to only 10\% of phishing URLs, which produces a separation of approximately 40\% that is consistent 
across both training and test sets. This consistency confirms that \textit{meta\_info\_present} is a reliable and stable contextual signal for post-hoc belief revision.

\subsection{Phase 2: NMR Belief Revision Results}
We used non-monotonic reasoning  to revise the initial classifier prediction. These predictions are treated as initial beliefs and fed as input to NMR layer.
\subsubsection{Instance-Level Belief Revision Analysis}
 Based on the domain specific rules, the reasoning layer selectively revises certain predictions, enabling belief updates. To illustrate the belief revision, two representative instances from the test sets are examined. \\Instance \texttt{url464}, (\url{http://www.allquests.com/question/2248296-1/}) is an actual legitimate URL. SVM, KNN and RF predicted as phishing, while DT predicted as legitimate.  Since, meta information is present, the URL length is 44 characters long, the \texttt{www} prefix is present and domain is not embedded in the page title. No defeater applies, so rule 1 fires and the three phishing predictions are revised to legitimate. All three revisions are correct.
 \\
 Instance \texttt{url2048} (\url{https://www.carmi.be/nl/home}) is also a legitimate URL. SVM and DT predicted as phishing while KNN and RF predicted as it as legitimate. Meta information is present, URL is 28 characters long, \texttt{www} prefix present, and domain not in title. Rule 1 again fired and both phishing predictions are revised to legitimate. Both revision are correct.

\subsubsection{Revised Predictions Count}

 Table \ref{tab:revised_count} presents the count of instances whose initial belief or prediction was revised by the NMR layer for each classifier. Overall, out of 9,144, classifier level predictions (2,286 test instances 
$\times$ 4 classifiers), 5 were revised by the NMR reasoning layer, representing a revision rate of 0.05\%.

\begin{table}[h]
\centering
\caption{Number of Revised Predictions 
by NMR Layer}
\label{tab:revised_count}
\small
\begin{tabular}{|l|c|}
\hline
\textbf{Classifier} & 
\textbf{Revised Predictions} \\
\hline
SVM & 2 \\
KNN & 1 \\
Decision Tree & 1 \\
Random Forest & 1 \\
\hline
\textbf{Total} & \textbf{5} \\
\hline
\end{tabular}
\end{table}

\subsubsection{Performance Comparison Before and After NMR}
Table \ref{tab:nmr_results} presents the complete performance comparison before and after NMR-based belief revision using the proposed 4-rule defeasible framework.  

\begin{table*}[h]
\centering
\caption{Performance After NMR-Based 
Belief Revision --- PHISHREV+ Framework}
\label{tab:nmr_results}
\small
\begin{tabular}{|l|c|c|c|c|c|c|c|c|c|}
\hline
\textbf{Classifier} & 
\multicolumn{4}{c|}{\textbf{Before NMR}} & 
\multicolumn{4}{c|}{\textbf{After NMR}} & 
\textbf{Revised} \\
\cline{2-9}
& \textbf{Acc(\%)} & 
\textbf{Prec} & 
\textbf{FP} & 
\textbf{FN} & 
\textbf{Acc(\%)} & 
\textbf{Prec} & 
\textbf{FP} & 
\textbf{FN} & \\
\hline
SVM & 
96.33 & 0.9641 & 41 & 43 & 
96.41 & 0.9658 & 39 & 43 & 2 \\
\hline
KNN & 
95.45 & 0.9585 & 47 & 57 & 
95.49 & 0.9594 & 46 & 57 & 1 \\
\hline
Decision Tree & 
93.70 & 0.9448 & 62 & 82 & 
93.74 & 0.9456 & 61 & 82 & 1 \\
\hline
Random Forest & 
95.80 & 0.9548 & 52 & 44 & 
95.84 & 0.9557 & 51 & 44 & 1 \\
\hline
\end{tabular}
\end{table*}

The proposed framework achieves consistent false positive reduction across all four classifiers - SVM((41$\rightarrow$39),KNN (47$\rightarrow$46), 
DT (62$\rightarrow$61), and RF (52$\rightarrow$51) while maintaining complete false negative parity with the base classifier, with zero false negative increase. Precision improves consistently across all classifiers following NMR revision, reflecting the reduction in false positives. Recall remains identical before and after NMR revision, for all classifiers, consistent with the zero false negative change.
The consistent false positive reduction across all classifiers directly addresses the problem of alert fatigue.\\
A McNemar test on paired predictions was used to compare ML-only and ML+NMR for each classifier. Let $b$ be the number of predictions that were wrong before revision and correct after it, and $c$ the number that were correct before and wrong after. Table~\ref{tab:mcnemar} reports the result. For every classifier, $c = 0$, so the NMR  layer corrected false positives and introduced no new errors. The direction of betterment is therefore consistent in favor of ML+NMR. However, only one or two predictions change per classifier, so the difference does not reach statistical significance on this test set(exact $p \geq 0.5$). Larger test sets, where more instances satisfy the revisions conditions are expected to give more corrections and greater statistical power.

\begin{table}[h]
\centering
\caption{McNemar test on paired predictions, ML-only vs ML+NMR (test set). $b$: wrong before and correct after revision. $c$: correct before and wrong after revision.}
\label{tab:mcnemar}
\small
\begin{tabular}{|l|c|c|c|}
\hline
\textbf{Classifier} & $\boldsymbol{b}$ & $\boldsymbol{c}$ & \textbf{Exact $p$} \\
\hline
SVM & 2 & 0 & 0.500 \\
KNN & 1 & 0 & 1.000 \\
DT  & 1 & 0 & 1.000 \\
RF  & 1 & 0 & 1.000 \\
\hline
\end{tabular}
\end{table}

\subsection{Comparison with Deterministic Post-Processing}\label{sec:ifelse}
To isolate the contribution of the ASP formalism from that of rule based system, the same four rules were implemented as a conventional deterministic if-else procedure and applied to the same classifier predictions. As shown in Table~\ref{tab:ifelse_baseline}, the if-else baseline and the ASP-based NMR layer produce identical FP and FN counts across all four classifiers and an instance level comparison confirms that both revise exactly the same predictions (\texttt{url464} for 
SVM, KNN and RF; \texttt{url2048} for SVM and DT).\\
\begin{table}[h]
\centering
\caption{Comparison of ML-only, Deterministic 
If-Else Post-Processing, and ASP-Based NMR 
(FP / FN)}
\label{tab:ifelse_baseline}
\small
\begin{tabular}{|l|c|c|c|}
\hline
\textbf{Classifier} & 
\textbf{ML-only} & 
\textbf{ML + If-Else} & 
\textbf{ML + ASP} \\
\hline
SVM & 41 / 43 & 39 / 43 & 39 / 43 \\
\hline
KNN & 47 / 57 & 46 / 57 & 46 / 57 \\
\hline
DT  & 62 / 82 & 61 / 82 & 61 / 82 \\
\hline
RF  & 52 / 44 & 51 / 44 & 51 / 44 \\
\hline
\end{tabular}
\end{table}
The value of the ASP lies not in accuracy, but in representation. Rules are declarative and order-independent, defaults and defeaters are explicit,  a unique answer set is guaranteed 
by stable model semantics~\cite{gelfond1988stable}, each revision has a formal derivation that supports XASP-based explanation 
(Section~\ref{sec:xasp}), and a new rule is added as a single rule rather than by editing procedural control flow. 

\subsection{Runtime Analysis}\label{sec:runtime}
To complement the theoretical analysis, execution times were measured on a personal computer. Runtime measurements report the median of five runs. 
As shown in table \ref{tab:runtime}, the complete NMR layer, comprising fact generation, grounding and solving, process all 9,144 classifier level decision in 0.313 seconds, of which Clingo grounding and solving account for only 0.062 seconds. Table \ref{tab:scaling}, reports grounding and solving time as the number of test instances increase from 500 to 2,286. The per-instance cost remains approximately constant at 0.027 ms. By comparison, retraining a single SVM or Random Forest model with fixed hyper-parameters requires 7.67 s and 2.88 s respectively, excluding the cost of hyperparameter search. 

\begin{table}[h]
\centering
\caption{Empirical runtime of the ML phase and the NMR layer 
(median of five runs)}
\label{tab:runtime}
\small
\begin{tabular}{|l|c|c|}
\hline
\textbf{Component} & 
\textbf{Training (s)} & 
\textbf{Inference (s)} \\
\hline
SVM & 7.665 & 0.4116 \\
\hline
KNN & 0.003 & 0.4534 \\
\hline
DT  & 0.126 & 0.0006 \\
\hline
RF  & 2.876 & 0.0610 \\
\hline
\multicolumn{3}{|l|}{\textbf{NMR layer} 
($n$ = 2,286; 9,144 decisions)} \\
\hline
Fact generation & \multicolumn{2}{c|}{0.2507} \\
\hline
Clingo grounding & \multicolumn{2}{c|}{0.0554} \\
\hline
Clingo solving & \multicolumn{2}{c|}{0.0071} \\
\hline
\textbf{Total NMR} & 
\multicolumn{2}{c|}{\textbf{0.3132}} \\
\hline
\end{tabular}
\end{table}

\begin{table}[h]
\centering
\caption{Scaling of ASP grounding and solving time with the 
number of test instances (median of five runs).Grounding instantiates the rules with the facts of each instance; 
solving computes the unique answer set of the resulting ground program.}
\label{tab:scaling}
\small
\begin{tabular}{|c|c|c|}
\hline
\textbf{$n$} & 
\textbf{Ground + Solve (s)} & 
\textbf{ms / instance} \\
\hline
500  & 0.0140 & 0.0281 \\
1000 & 0.0267 & 0.0267 \\
1500 & 0.0423 & 0.0282 \\
2000 & 0.0535 & 0.0267 \\
2286 & 0.0614 & 0.0268 \\
\hline
\end{tabular}
\end{table}

\section{Ablation Study}\label{sec:ablation}
To evaluate the individual and collective contribution of each defeasible rule to the overall PHISHREV+ framework, a comprehensive ablation study is conducted by progressively adding rules to the NMR knowledge base and evaluating the resulting FP and FN counts across all four classifiers. Total 7 rule configurations are evaluated, ranging from single rule baseline to the proposed 4-rule framework. The ablation study addresses the \textbf{RQ3} that whether the hierarchical interaction between multiple defeasible rules achieves an optimal balance between false positive reduction and false negative preservation.
\subsection{Rule Configurations}

The following rule configurations are 
evaluated:

\begin{itemize}
\item \textbf{Rule 1 only:} Base revision rule\\
Revise phishing to legitimate if meta= yes
\item \textbf{Rule 1+2:}\\
Base revision with URL length suppression 
(length\_url $>$ 63).

\item \textbf{Combination A (Rule 1+3):} \\
Base revision with www suppression (nb\_www = 0).

\item \textbf{Combination B (Rule 1+2+3):}\\ 
Base revision with URL length and www suppression.

\item \textbf{Combination C (Rule 1+3+4):}\\ 
Base revision with www and domain-title suppression.

\item \textbf{Combination E (Rule 1+2+4):}\\ 
Base revision with URL length and domain-title defeaters

\item \textbf{Proposed (Rule 1+2+3+4):}\\ 
Complete 4-rule framework with all suppression conditions.
\end{itemize}

\subsection{Ablation Results}
The results in table \ref{tab:ablation}, presents the complete ablation results across all four classifiers for each rule configuration. 

\begin{table*}[h]
\centering
\caption{Ablation Study: FP and FN 
across all classifiers for each 
rule configuration}
\label{tab:ablation}
\small
\resizebox{\textwidth}{!}{%
\begin{tabular}{|l|cc|cc|cc|cc|c|c|}
\hline
\textbf{Configuration} & 
\multicolumn{2}{c|}{\textbf{SVM}} & 
\multicolumn{2}{c|}{\textbf{KNN}} & 
\multicolumn{2}{c|}{\textbf{DT}} & 
\multicolumn{2}{c|}{\textbf{RF}} & 
\textbf{Total} & 
\textbf{RF FN} \\
& \textbf{FP} & \textbf{FN} & 
\textbf{FP} & \textbf{FN} & 
\textbf{FP} & \textbf{FN} & 
\textbf{FP} & \textbf{FN} & 
\textbf{Revised} & 
\textbf{Change} \\
\hline
Before NMR & 
41 & 43 & 47 & 57 & 62 & 82 & 52 & 44 & 
--- & --- \\
\hline
Rule 1 only & 
30 & 145 & 33 & 159 & 42 & 177 & 38 & 146 & 
460 & +102 \\
\hline
Rule 1+2 & 
33 & 100 & 34 & 115 & 49 & 135 & 41 & 101 & 
270 & +57 \\
\hline
Comb A (Rule 1+3) & 
35 & 68 & 42 & 78 & 53 & 106 & 46 & 68 & 
120 & +24 \\
\hline
Comb B (Rule 1+2+3) & 
37 & 55 & 43 & 67 & 57 & 93 & 48 & 56 & 
62 & +12 \\
\hline
Comb C (Rule 1+3+4) & 
39 & 45 & 46 & 59 & 61 & 84 & 51 & 46 & 
13 & +2 \\
\hline
Comb E (Rule 1+2+4) & 
39 & 47 & 41 & 61 & 58 & 86 & 49 & 48 & 
31 & +4 \\
\hline
\textbf{Proposed (1+2+3+4)} & 
\textbf{39} & \textbf{43} & 
\textbf{46} & \textbf{57} & 
\textbf{61} & \textbf{82} & 
\textbf{51} & \textbf{44} & 
\textbf{5} & \textbf{0} \\
\hline
\end{tabular}}
\end{table*}
\subsection{Analysis}
In this subsection, we provide detail analysis of the outcome of each combination of rule applied.
\subsubsection{Rule 1 Alone}
When rule 1 is applied only, the base revision rule that revises all phishing prediction with meta=YES to legitimate, the framework achieves the highest false positive reduction across all classifiers (SVM: 41$\rightarrow$30, KNN: 47$\rightarrow$33, DT: 62$\rightarrow$42, RF: 52$\rightarrow$38) as presented in table \ref{tab:ablation}. However, this aggressive revision comes at a significant cost: false negatives increases substantially across all classifiers, with RF false negatives rising from 44 to 146, DT false negatives rising from 82 to 177, KNN false negatives rising from 57 to 159 and SVM false negatives rising from 43 to 145. This demonstrate that a single rule without any defeaters is insufficient for balanced belief revision that is it cannot distinguish genuinely legitimate pages from phishing pages.     
\subsubsection{Rule 1+2: URL Length Defeater}
Addition of rule 2 which defeats revision when URL length exceeds $\tau = 63$ characters, significantly reduces the false negative increases(RF: 146$\rightarrow$101, DT 177$\rightarrow$135, KNN 159$\rightarrow$115, SVM 145$\rightarrow$100)while maintaining good false positive reduction. 
\subsubsection{Combinations A, B, C and E}
Combinations A, B, C and E as shown in table \ref{tab:ablation} progressively add defeater rules, showing a consistent trend: as more defeater rule conditions are introduced, the false negative increase is progressively controlled at the cost of reduced false positives reductions. For RF, the false negative increase falls to $+24$ (A), $+12$ (B), $+4$ (E) and $+2$ (C). Combination C comes closest to the proposed framework but still introduces two additional false negatives.These results confirm that 
each defeater rule contributes a complementary signal that collectively tightens the revision conditions.

\subsubsection{Proposed 4-Rule Framework}
The proposed 4-rule framework (Rule 1+2+3+4) achieves the optimal configuration across all four classifiers. 

\begin{itemize}
\item \textbf{SVM:} FP reduces from 41 to 39 while FN remains unchanged at 43 (zero increase).
\item \textbf{KNN:} FP reduces from 47 to 46 while FN remains unchanged at 57 (zero increase).
\item \textbf{DT:} FP reduces from 62 to 61 while FN remains unchanged at 82 (zero increase).
\item \textbf{RF:} FP reduces from 52 to 51 while FN remains unchanged at 44 (zero increase).
\end{itemize}
This is the only configuration that achieves simultaneous false positive reduction and complete false negative parity across all four classifiers. It revises 5 classifier level 
predictions (5/9,144 = 0.05\%), all of which are correct. 

\subsection{Narrowing of the revision search space}
\begin{figure}[h]
\centering
\includegraphics[width=0.85\textwidth]{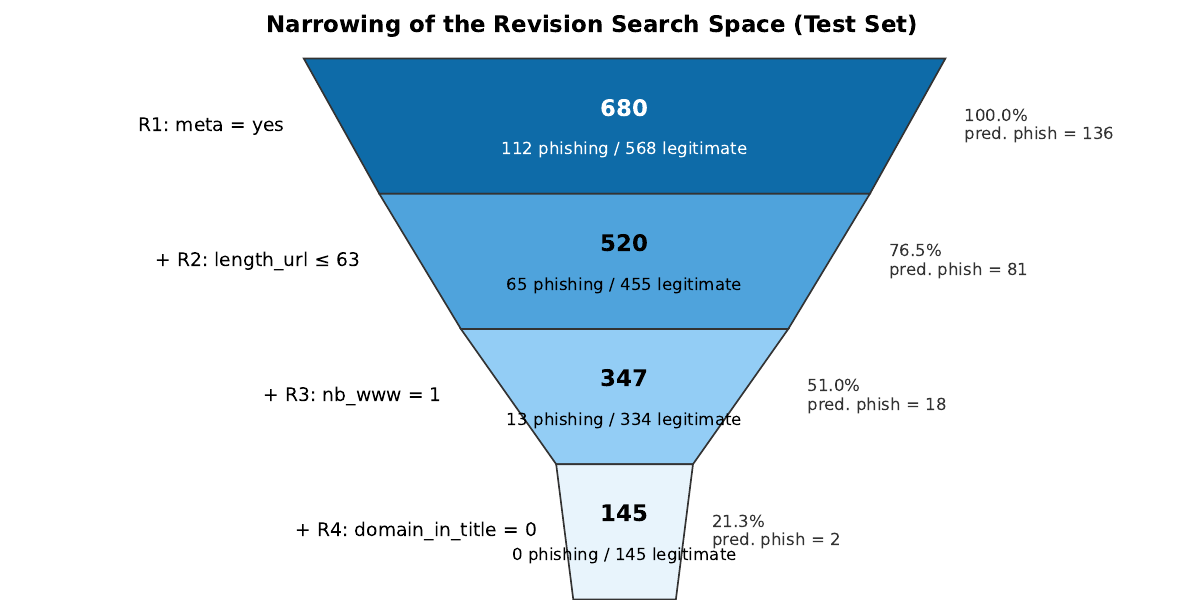}
\caption{Narrowing of the revision search space on the test set. 
Each stage retains only the instances not blocked by the 
corresponding defeater rule; band width is proportional to the 
number of instances. Rule~1 admits 680 instances with meta 
information present (112 phishing). Successive defeaters --- 
URL length (Rule~2), \texttt{www} prefix (Rule~3) and domain in 
title (Rule~4) --- reduce this set to 520, 347 and finally 145 
instances (21.3\% of the initial set), none of which is phishing. 
Of these, only two were predicted as phishing by any classifier, 
and these are exactly the instances revised by the NMR layer.}
\label{fig:funnel}
\end{figure}

Table \ref{tab:funnel} and Figure~\ref{fig:funnel}, show how the defeater rules progressively narrow the set of test instances on which the default revision rule can act. Rule 1, admits all 680 instances with meta information present, of which 112 are phishing. Requiring a URL length of at most $\tau = 63$ characters reduces this set to 520 samples (65 phishing), requiring the \texttt{www} prefix (Rule 3) reduces it to 347 (13 phishing). And at last rule 4, leaves only 145 instances, none of which is phishing. 
Consequently, no true phishing instances, remains on which rule 1 can fire, which explains zero false negative increase observed across all classifiers.
Of these 145 legitimate instances, only two instances 
--- \texttt{url464} and \texttt{url2048}, were predicted as phishing by at least one classifier, and these are precisely the instances revised by the NMR layer.

\begin{table}[h]
\centering
\caption{Narrowing of the revision search space on the test set. Each row retains the instances that pass all preceding conditions, i.e., those 
not blocked by any defeater.}
\label{tab:funnel}
\small
\begin{tabular}{|l|c|c|c|p{1.2cm}|p{1.0cm}|}
\hline
\textbf{Stage} & 
\textbf{URLs} & 
\textbf{Phish.} & 
\textbf{Legit.} & 
\textbf{Pred. phish.$^{a}$} & 
\textbf{Pred. legit.$^{b}$} \\
\hline
R1: meta = yes & 680 & 112 & 568 & 136 & 544 \\
\hline
+ R2: length\_url $\leq$ 63 & 520 & 65 & 455 & 81 & 439 \\
\hline
+ R3: nb\_www = 1 & 347 & 13 & 334 & 18 & 329 \\
\hline
+ R4: domain\_in\_title = 0 & 
\textbf{145} & \textbf{0} & \textbf{145} & 
\textbf{2} & \textbf{143} \\
\hline
\multicolumn{6}{l}{\footnotesize $^{a}$Predicted phishing by at least one classifier.} \\
\multicolumn{6}{l}{\footnotesize $^{b}$Predicted legitimate by all four classifiers.} \\
\end{tabular}
\end{table}

\subsection{Progressive Refinement of Revisions}
Table~\ref{tab:funnel} counts unique 
instances, where as Table \ref{tab:cascade}, counts classifier-level revised predictions (test instances $\times$ classifiers), summed over all four classifier. Rule 1, when applied alone, revises 460 predictions with 12.8\% precision while rule 1+ rule 2+rule 3+rule 4 revises 5 predictions, all of which are correct. The narrowing comes at a cost: the number of corrected false positives also falls, from 59 to 5, reflecting the conservative design of the framework, which prioritises zero false negative increase over the magnitude of false positive reduction. 

\subsection{Implications of the Ablation Study}
The ablation study presented supports following three observations. 
\begin{enumerate}

\item \textbf{No single rule is sufficient:}
no individual rule or partial combinations achieves simultaneous false positives reduction and zero false negative increase. Each defeater removes a distinct subset of incorrect revisions admitted by the default rule in table~\ref{tab:funnel} 
, and only the full combination eliminates them all.
\item \textbf{Defeaters are complementary:} the rules act on different contextual signals like URL length, \texttt{www} prefix, domain in title etc. and their effects accumulate rather than overlap, as shown by the progressive decrease in false negative increase 
for RF from $+102$ (Rule 1) to $+57$ (Rules 1-2), $+12$ (Rules 1-3) and $0$ (Rules 1-4) in table \ref{tab:ablation}.
\item \textbf{Incremental knowledge incorporation:} each configuration was obtained by adding defeasible rules to the ASP knowledge base and re-evaluating only the reasoning layer over the same, unchanged classifier predictions, no classifier was retrained at any stage. Each update required only re-grounding and re-solving the program, which completes in approximately 0.06~s for all 9144 classifier level decisions (Table~\ref{tab:runtime}) and scales linearly with the number of instances (Table~\ref{tab:scaling}). This proves empirical support for \textbf{RQ4}.
\end{enumerate}

\begin{table}[h]
\centering
\caption{Progressive narrowing of revisions 
as defeater rules are added. Counts are 
classifier-level predictions (test instances 
$\times$ classifiers), summed over all four 
classifiers.}
\label{tab:cascade}
\small
\begin{tabular}{|l|c|c|c|c|}
\hline
\textbf{Rule set} & 
\textbf{Revised predictions} & 
\textbf{Correct} & 
\textbf{Incorrect} & 
\textbf{Precision} \\
\hline
R1 & 460 & 59 & 401 & 12.8\% \\
\hline
R1 + R2 & 270 & 45 & 225 & 16.7\% \\
\hline
R1 + R2 + R3 & 62 & 17 & 45 & 27.4\% \\
\hline
\textbf{R1 + R2 + R3 + R4} & 
\textbf{5} & \textbf{5} & 
\textbf{0} & \textbf{100\%} \\
\hline
\end{tabular}
\end{table}

\subsection{Need of Answer Set Programming}
The four rules interact hierarchically. Rule 1 is a default that revises a phishing prediction to legitimate, while rules 2-4 are defeaters that block this default when conflicting contextual evidence is present. This pattern, in which a default conclusion holds unless an exception is established, is the canonical form of non-monotonic reasoning and ASP expresses it directly through negation as failure (\texttt{not}): 
Rule~1 fires only if none of the defeating 
atoms can be derived.\\
The same hierarchy can be implemented procedurally, and Section \ref{sec:ifelse}
confirms that an equivalent if-else procedure produces identical predictions. The motivation for ASP is therefore not predictive superiority but the properties of the representation, summerised in table \ref{tab:asp_vs_ifelse}.
\begin{table}[h]
\centering
\caption{Qualitative comparison of deterministic 
if-else post-processing and ASP-based 
non-monotonic reasoning for the proposed 
rule set}
\label{tab:asp_vs_ifelse}
\small
\begin{tabular}{|p{3cm}|p{4.8cm}|p{4.8cm}|}
\hline
\textbf{Aspect} & 
\textbf{If-else procedure} & 
\textbf{ASP knowledge base} \\
\hline
Predictions & 
Identical to ASP (Table~\ref{tab:ifelse_baseline}) & 
Identical to if-else \\
\hline
Rule priority & 
Implicit in nesting order & 
Explicit via default and defeater rules \\
\hline
Adding a rule & 
Edit control flow at the correct position & 
Append one rule \\
\hline
Semantics & 
Defined by program execution & 
Stable model semantics; unique answer set \\
\hline
Explanation & 
Returns a label only & 
Formal derivation per revision (XASP) \\
\hline
Knowledge vs.\ code & 
Rules embedded in code & 
Rules separate from ML pipeline \\
\hline
\end{tabular}
\end{table}

\begin{enumerate}
\item \textbf{Declarative rules:} The priority between the default and its defeaters is stated explicitly in the rule bodies rather than encoded implicitly in nesting order of conditional statements.
\item \textbf{Well-defined semantics:} The program is stratified and non-disjunctive, so a stable model semantics \cite{gelfond1988stable} guarantees a unique answer set. 
\item \textbf{Explainability:} Every revised belief has a formal derivation from facts and rules, which enables the rule level explanations presented  in Section~\ref{sec:xasp}.
\item \textbf{Extensibility:} The knowledge base is separated from machine learning pipeline, so a new defeater is incorporated by appending a single rule, without modifying procedural control flow or retraining classifier.
\end{enumerate}

\section{Comparison with state of art}
\label{stateofartworks}
Table \ref{tab:sota} compares PHISHREV+ against existing phishing website detection studies that used the same benchmark. Prior works, focus exclusively on maximizing classification accuracy. They report overall accuracy but not false positive or false negatives. In these studies, the prediction of the classifier is final. The decision cannot be revised at inference time and new knowledge requires retraining the model. The authors of \cite{phishrev2026}, proposed post-hoc non-monotonic revision of classifier prediction. Their work uses a single rule that revises a phishing prediction to legitimate when meta information is present. That single rule corresponds to the ``Rule~1 only'' configuration of the ablation study (Table~\ref{tab:ablation}). On the current work, it reduces false positive but sharply increases false negatives. Our proposed work, PHISHREV+ extends this work with three defeaters derived from the training set, which removes this cost of increasing false negatives. Accuracy of the proposed work is comparable to prior work but its contribution is different capability: initial predictions can be revised and explained at inference time, and new knowledge is added without retraining the classifier. 
\begin{table*}[h]
\centering
\caption{Comparison with prior work on the same benchmark dataset. n.r.\ = not reported in the original paper.}
\label{tab:sota}
\small
\begin{tabular}{|p{3cm}|p{1.9cm}|p{1.5cm}|p{1.8cm}|p{3cm}|p{2.5cm}|}
\hline
\textbf{Aspect} & \textbf{Hannousse et al.~\cite{hannousse2021towards}} & \textbf{Adane et al.~\cite{adane2023single}} & \textbf{Shafin~\cite{shafin2025explainable}} & \textbf{PHISHREV~\cite{phishrev2026}} & \textbf{PHISHREV+} \\
\hline
RF accuracy (\%) & 96.61 & 96.54 & 96.76 & 95.89 & 95.80 / 95.84$^{a}$ \\
\hline
FP / FN reported & n.r. & n.r. & n.r. & FP only$^{b}$ & Yes \\
\hline
Post-hoc revision of predictions & No & No & No & Yes & Yes \\
\hline
Number of revision rules & -- & -- & -- & 1 & 4 (1 default, 3 defeaters) \\
\hline
New knowledge needs retraining & Yes & Yes & Yes & No & No \\
\hline
Explanation & None & None & Feature-level & None & Rule-level (XASP) \\
\hline
\end{tabular}

\vspace{2pt}
\raggedright\footnotesize $^{a}$Before / after NMR-based belief revision.\\
$^{b}$As reported in the original paper. FN was not reported.
\end{table*}

\section{Explainability Analysis via XASP} \label{sec:xasp}
Explainability is a critical requirement in cybersecurity applications where classification decisions must be transparent and auditable. However, existing explanations made by SHAP or LIME, do not explain the post-hoc belief revision decision made by the NMR reasoning layer. Specifically, they do not provide insights to which defeasible rules fired, which were defeated, and why the initial belief was formally retracted. This section presents an instance-level explainability analysis of the NMR belief revision using eXplainable ASP (XASP). This provides formal, rule-level explanations through a directed acyclic graph(DAG)representation of the ASP derivation chain \cite{alviano_xasp}.  XASP identifies (i). which defeasible rules were evaluated, (ii). which rules were defeated and why, (iii). which rule fired and what condition, (iv).how the initial belief was formally was retracted to produce the final belief. This three level explainability, fact level, rule level and derivation level, provides a completes and auditable justification for each belief revision decision. 

\subsection{XASP Explanation Framework}
For each revised instance, the XASP analysis traces the complete derivation chain of the NMR reasoning layer across five levels.

\begin{enumerate}
\item \textbf{Facts (Layer 1):} The ground ASP facts for the instance presents initial classifier prediction, meta tag availability, URL length, www prefix, and domain-title signal.  \item \textbf{Defeaters Evaluated (Layer 2):} Evaluation of rules 2,3 and 4, whether each defeating condition is satisfied. A defeated rule is shown with a dashed connector and marked as not fired.  
\item \textbf{Rule Fired (Layer 3):} The defeasible  rule that fires based on the fact evaluation, rule 1 fires when all three defeating conditions are absent. 
\item \textbf{Belief Revision (Layer 4):} The formal retraction of initial belief $\mathcal{B}_0$ and derivation of final belief in $\mathcal{B}_0$ time. 
\item \textbf{Outcome (Layer 5):} Whether the revision is correct, verified against the true label.  
\end{enumerate}
The XASP explanation DAGs were generated using the xasp library~\cite{alviano_xasp} and visualized programmatically from library's explanation DAG output. \textcolor{blue}{The full visualization script will be made available with the other code upon acceptance.}

\subsection{XASP Analysis --- Instance url464}
\begin{figure*}[!t]\centering\includegraphics[width=14cm, height=12cm, keepaspectratio]{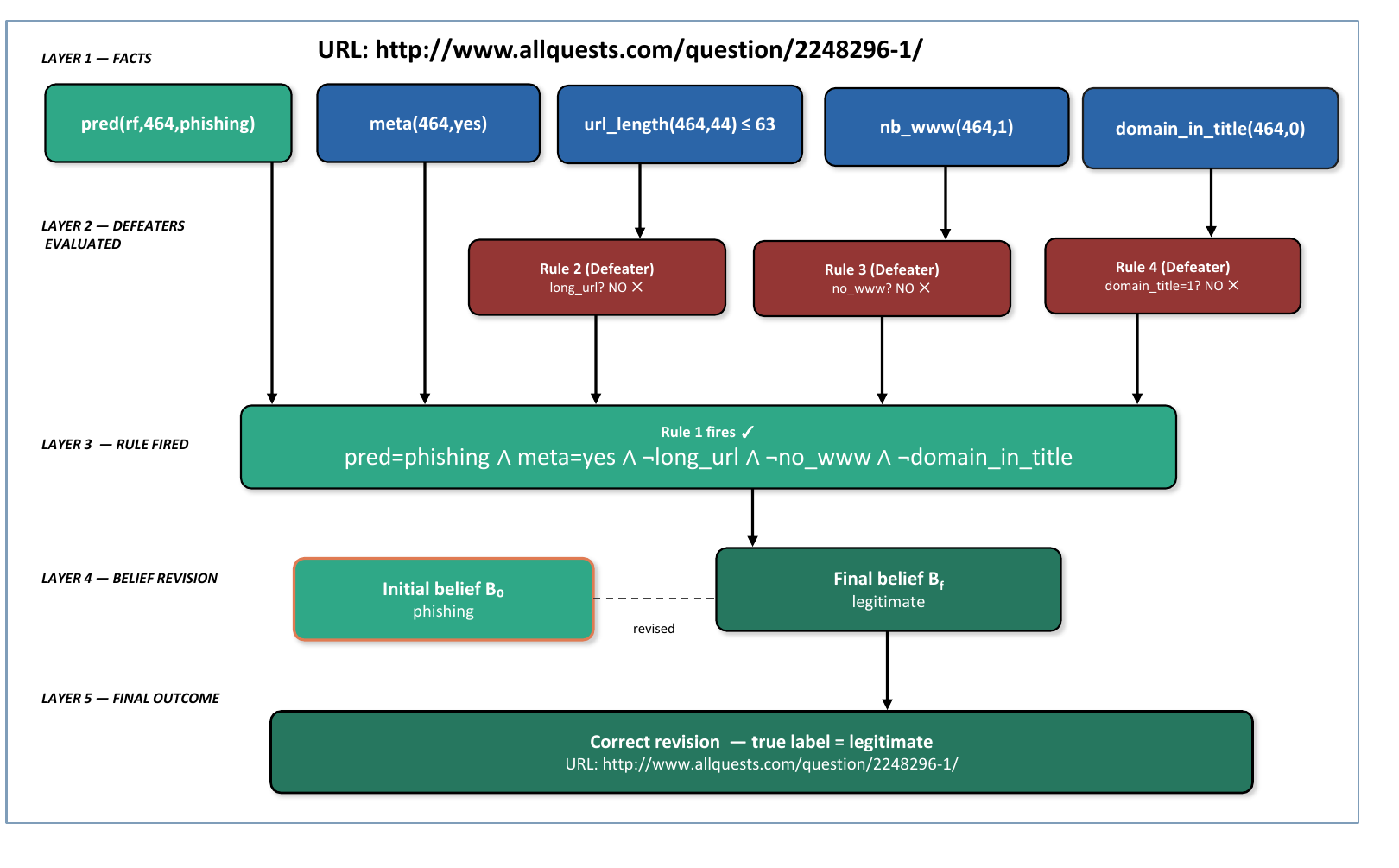}\caption{XASP explanation DAG for instance \texttt{url464} 
generated using the xasp library~\cite{alviano_xasp}. Teal nodes indicate true facts, fired rules and initial belief $\mathcal{B}_0$; blue nodes indicate contextual facts, brown nodes denote defeated rules, dark green indicates the final belief $\mathcal{B}_f$. Rule 1 fires because all three defeating conditions are absent, formally retracting the initial phishing belief and revising to legitimate, thus  correctly reducing a false positive. The same derivation applies to the SVM and KNN predictions for this instance. DT predicted legitimate, so no revision was needed.}
\label{fig:xasp_464}
\end{figure*}

Figure~\ref{fig:xasp_464} presents the XASP derivation DAG for instance \texttt{url464} (\url{http://www.allquests.com/question/2248296-1/}), a legitimate URL incorrectly predicted as phishing by SVM, KNN and RF. The figure shows the RF derivation.  

\section{Discussion}\label{discussion}
This section discusses the findings of the proposed framework, interprets the experimental results and places the limitations and broader implications of the work.

\subsection{FP Reduction and FN Parity}
The primary objective of PHISHREV+ is not accuracy maximization but the demonstration that contextual evidence can be formally incorporated into phishing classification decisions through defeasible reasoning in $\mathcal{O}(n)$ time. The experimental results confirms that the proposed framework achieves consistent false positive reduction across all four classifier-SVM(41$\rightarrow$39), KNN 
(47$\rightarrow$46), DT 
(62$\rightarrow$61), and RF 
(52$\rightarrow$51) while maintaining complete false negative parity with the base classifier. 

\subsection{Addressing the Research Questions}
The EDA analysis conducted on training set confirms that all four contextual features --- 
\textit{meta\_info\_present} (40.72\% separation; 
$\chi^2 = 1790.46$, OR = 9.26), length\_url (mean 63.22 vs 
45.47 characters, yielding $\tau = 63$; Mann--Whitney 
$p = 1.23 \times 10^{-14}$, small effect $r = 0.146$), \textit{nb\_www} (52.8\% separation, the strongest binary signal; $\chi^2 = 336.96$, OR = 10.48) and \textit{domain\_in\_title} 
(25.8\% separation; $\chi^2 = 16.95$, OR = 3.62) exhibit statistically meaningful distributional differences that motivate the defeasible rule design. All four reject the null hypothesis of independence at $\alpha = 0.05$, and together they motivate the defeasible rule design, which was frozen before evaluation on untouched test set.\\
The results of the proposed framework confirms this: false positive decrease for all four classifiers (SVM 41$\rightarrow$39, KNN 
47$\rightarrow$46, DT 62$\rightarrow$61, RF 52$\rightarrow$51) while false negatives remain unchanged, with 5 of 9,144 classifier-level predictions revised(0.05\%).

The ablation study shows that no single rule or partial combination achieves false positive reduction without increasing false negatives. Only the full 4-rule hierarchy does so, with the search space narrowing from 680 to 145 instances, none of which is phishing. A deterministic if-else implementation of the same rules produces identical predictions, so the predictive gain derives from the EDA-derived rules. The contribution of ASP lies in its declarative, order-independent representation of defaults and defeaters, its well defined stable model semantics and its support for formal rule-level explanation through XASP.  \\
This is supported both theoretically, by the complexity analysis of the stratified, non-disjunctive rule fragment and empirically: the ablation study incorporates successive defeater rules without retraining any classifier, and grounding and solving time grows linearly with the number of instances, at approximately 0.027~ms per instance.
\subsection{Domain-Title Signal}
A key finding of the EDA analysis is counter-intuitive behaviour of \textit{domain\_in\_title}. In the training-set subset that passes rules 2 and 3,  82.6\% of phishing instances (57 of 69) embed the 
domain in the page title, compared with 56.8\% of legitimate instances (787 of 1,386; OR = 3.62). This pattern is consistent with the hypothesis that phishing pages may deliberately reproduce characteristics of legitimate pages, such as meta-information and domain-bearing title, although the dataset does not directly establish attacker intent. The association motivates rule 4, as a defeater: the ablation study shows that omitting it (Combination~B) leaves a false negative increase of $+12$ for RF, which is removed in rule 4. This illustrates how EDA can surface non-obvious contextual signals that are then encoded explicitly as defeasible rules.
\subsection{Selective and Evidence-Driven Revision}
The proposed framework revises only 5 instances out of 9,144 
classifier-level predictions that is a revision rate of 0.05\%. This selectivity is a direct consequence of the joint satisfaction requirement of rule 1 and it's three defeaters as shown in table \ref{tab:funnel}. Only 145 test instances satisfy all conditions, none of which is phishing. All 5 revisions are correct. The system corrects false positives and does not include false negatives, which directly address alert fatigue\cite{ban2023breaking}.

\subsection{Limitation}
Despite the promising results, the proposed PHISHREV+ framework has several limitations. The framework is evaluated on a single dataset and the generalizability of the EDA derived rules to datasets from different time periods remains to be validated. The small number of revised predictions also limits the statistical power of the paired comparison with the base classifiers.
The primary revision trigger \textit{meta\_info\_present} is a single binary signal that may lose effectiveness as adversarial pages increasingly embed meta tags to evade detection. Furthermore, the EDA derived threshold $\tau$ = 63 
for \textit{length\_url} may need recalibration when applied to datasets with different URL length distributions. Finally, the conservative nature of the 4-rule framework reflected in the 0.05\% revision rate, may miss legitimate revision opportunities  where contextual evidence is present but one defeating condition is simultaneously satisfied. \\
The proposed framework has broader implications beyond phishing detection. The general principle of combining statistical ML predictions with EDA-derived defeasible rules in an ASP based NMR layer is applicable to any classification problem where contextual evidence is available at inference time and false positive reduction is a primary objective. Examples include spam detection, intrusion detection and medical diagnosis where false positives carry significant operational costs. The $\mathcal{O}(n)$ knowledge update property further makes the framework suitable for dynamic environments where new domain knowledge must be incorporated rapidly without model retraining. 
\section{Conclusion and Future Scope}\label{conclusion}
This work presents PHISHREV+, a hybrid framework that integrates supervised machine learning classifiers with an ASP-based non-monotonic reasoning layer for context aware phishing website classification. Unlike existing approaches, that rely solely on statistical decision functions with fixed decision boundaries, PHISHREV+ introduces a post-hoc belief revision mechanism that formally revises initial classifier predictions using four defeasible rules encoded within an ASP-knowledge base. \\
The EDA-analysis identifies four contextual features namely,  \textit{meta\_info\_present}, \textit{length\_url}, \textit{nb\_www}, and \textit{domain\_in\_title}, exhibiting statistically meaningful distribution differences between phishing and legitimate URLs, with separations ranging from 11.1\% 
(\textit{length\_url}) to 52.8\% (\textit{nb\_www}). These findings directly motivate the hierarchical defeasible rule design, where rule 1 initiates belief revision and rules 2-4 act as defeaters that formally defeat rule 1 under conflicting evidence. The proposed method, achieves consistent false positive reduction across all four classifiers, SVM (41$\rightarrow$39), KNN (47$\rightarrow$46), DT (62$\rightarrow$61), and RF (52$\rightarrow$51), while maintaining complete false negative parity with base classifier, with zero false negative increase. A comprehensive ablation study confirms that no subset of rules achieves this balance on its own. All four rules are needed.\\   
The XASP analysis provides formal, rule-level explanations of the NMR belief revision process for two representative URLs \texttt{url464} and \texttt{url2048}. A key practical advantage of PHISHREV+ is that new domain knowledge can be incorporated directly into the ASP knowledge base in $\mathcal{O}(n)$ time, eliminating the need for costly model retraining. The framework performs selective evidence driven belief revision. It revises 
5 of 9,144 classifier-level predictions with 100\% revision precision.\\
Future work will consider in three directions. First, the framework will be extended to multiple phishing datasets spanning over different time periods to analyse temporal feature drift and derive drift informed defeasible rules. Secondly, the knowledge base will be enriched with additional contextual signals like domain age, SSL certificate characteristic etc. to further strengthen the reasoning layer.  Third, the proposed post-hoc defeasible machine learning paradigm will be evaluated across diverse application domains beyond phishing detection like spam classification, intrusion detection and medical diagnosis where false positive carry significant operational costs and contextual evidence is available at inference time.     

 \section{Data Availability}
The notebook file, .lp files, altered instances files will be made available upon acceptance. 


\bibliographystyle{elsarticle-num}
\bibliography{Eswa_references.bib}

@inproceedings{skula2024url,
  title={URL and domain obfuscation techniques-prevalence and trends observed on phishing data},
  author={Skula, Ivan and Kvet, Michal},
  booktitle={2024 IEEE 22nd World Symposium on Applied Machine Intelligence and Informatics (SAMI)},
  pages={000283--000290},
  year={2024},
  organization={IEEE}
}

@article{hannousse2021web,
  title={Web page phishing detection},
  author={Hannousse, Abdelhakim and Yahiouche, Salima},
  journal={Mendeley Data},
  volume={3},
  pages={2021},
  year={2021}
}

@article{mosa2023machine,
  title={Machine learning techniques for detecting phishing URL attacks},
  author={Mosa, Diana T and Shams, Mahmoud Y and Abohany, Amr A and El-kenawy, El-Sayed M and Thabet, M},
  journal={Computers, Materials \& Continua},
  volume={75},
  number={1},
  pages={1271--1290},
  year={2023},
  publisher={Tech Science Press}
}

@misc{apwg2026q2,
  author       = {{Anti-Phishing Working Group}},
  title        = {Phishing Activity Trends Report, 2nd Quarter 2026},
  year         = {2026},
  month        = aug,
  howpublished = {\url{https://docs.apwg.org/reports/apwg_trends_report_q2_2026.pdf}},
  note         = {Published 28 August 2026. Accessed: 15 October 2026}
}

@misc{phishrev2026,
  author        = {Sen, Mainak and Ray, Kumar Sankar and Chakrabarti, Amlan},
  title         = {{PHISHREV}: A Hybrid Machine Learning and Post-Hoc Non-monotonic Reasoning Framework for Context-Aware Phishing Website Classification},
  year          = {2026},
  eprint        = {2604.25512v1},
  archivePrefix = {arXiv},
  primaryClass  = {cs.AI},
  note          = {Version 1},
  url           = {https://arxiv.org/abs/2604.25512v1}
}

@article{afzal2021urldeepdetect,
  title={Urldeepdetect: A deep learning approach for detecting malicious urls using semantic vector models},
  author={Afzal, Sara and Asim, Muhammad and Javed, Abdul Rehman and Beg, Mirza Omer and Baker, Thar},
  journal={Journal of Network and Systems Management},
  volume={29},
  number={3},
  pages={21},
  year={2021},
  publisher={Springer}
}

@article{akccam2025development,
  title={Development of BiLSTM deep learning model to detect URL-based phishing attacks},
  author={Ak{\c{c}}am, {\"O}znur {\c{S}}ifa and Tekerek, Adem and Tekerek, Mehmet},
  journal={Computers and Electrical Engineering},
  volume={123},
  pages={110212},
  year={2025},
  publisher={Elsevier}
}

@article{alshehri2022character,
  title={Character-level word encoding deep learning model for combating cyber threats in phishing URL detection},
  author={Alshehri, Mohammed and Abugabah, Ahed and Algarni, Abdullah and Almotairi, Sultan},
  journal={Computers and Electrical Engineering},
  volume={100},
  pages={107868},
  year={2022},
  publisher={Elsevier}
}

@article{choudhary2023machine,
  title={A machine learning approach for phishing attack detection},
  author={Choudhary, Tarun and Mhapankar, Siddhesh and Bhddha, Rohit and Kharuk, Ashish and Patil, Rohini},
  journal={Journal of artificial intelligence and technology},
  volume={3},
  number={3},
  pages={108--113},
  year={2023}
}

@incollection{reiter1988nonmonotonic,
  title={Nonmonotonic reasoning},
  author={Reiter, Raymond},
  booktitle={Exploring artificial intelligence},
  pages={439--481},
  year={1988},
  publisher={Elsevier}
}

@article{asiri2024phishingrtds,
  title={PhishingRTDS: A real-time detection system for phishing attacks using a Deep Learning model},
  author={Asiri, Sultan and Xiao, Yang and Alzahrani, Saleh and Li, Tieshan},
  journal={Computers \& Security},
  volume={141},
  pages={103843},
  year={2024},
  publisher={Elsevier}
}

@article{xiao2021phishing,
  title={Phishing websites detection via CNN and multi-head self-attention on imbalanced datasets},
  author={Xiao, Xi and Xiao, Wentao and Zhang, Dianyan and Zhang, Bin and Hu, Guangwu and Li, Qing and Xia, Shutao},
  journal={Computers \& Security},
  volume={108},
  pages={102372},
  year={2021},
  publisher={Elsevier}
}

@article{gupta2024optimized,
  title={Optimized phishing detection with recurrent neural network and whale optimizer algorithm},
  author={Gupta, Brij Bhooshan and Gaurav, Akshat and Attar, Razaz Waheeb and Arya, Varsha and Alhomoud, Ahmed and Chui, Kwok Tai},
  journal={Computers, Materials and Continua},
  volume={80},
  number={3},
  pages={4895--4916},
  year={2024},
  publisher={Elsevier}
}

@article{heiding2024devising,
  author = {Heiding, Fredrik and 
            Schneier, Bruce and 
            Vishwanath, Arun and 
            Bernstein, Jeremy and 
            Park, Peter S.},
  title = {Devising and Detecting Phishing 
           Emails Using Large Language Models},
  journal = {IEEE Access},
  volume = {12},
  pages = {42131--42146},
  year = {2024},
  doi = {10.1109/ACCESS.2024.3375882}
}

@inproceedings{gelfond1988stable,
  author    = {Gelfond, Michael and 
               Lifschitz, Vladimir},
  title     = {The Stable Model Semantics 
               for Logic Programming},
  booktitle = {Proceedings of the 5th 
               International Conference 
               and Symposium on Logic 
               Programming},
  pages     = {1070--1080},
  year      = {1988},
  publisher = {MIT Press},
  address   = {Cambridge, MA, USA}
}

@inproceedings{madzime2026defeasible,
  author    = {Madzime, Ruvarashe},
  title     = {Toward a Defeasible 
               Semantics for Symbolic 
               Classifiers},
  booktitle = {Doctoral Consortium of 
               the 23rd International 
               Conference on Principles 
               of Knowledge Representation 
               and Reasoning (KR 2026 DC)},
  year      = {2026},
  url       = {https://ceur-ws.org/
               Vol-4218/paper3.pdf},
  note      = {CEUR Workshop Proceedings, 
               Vol. 4218}
}

@inproceedings{chataut2024llm,
  author = {Chataut, Robin and 
            Gyawali, Prajwol Kumar and 
            Usman, Yamen},
  title = {Can {AI} Keep You Safe? A Study 
           of Large Language Models for 
           Phishing Detection},
  booktitle = {Proceedings of the 2024 IEEE 
               14th Annual Computing and 
               Communication Workshop and 
               Conference (CCWC)},
  pages = {548--554},
  year = {2024},
  doi = {10.1109/CCWC60891.2024.10427626}
}

@misc{alviano_xasp,
  author       = {Alviano, Mario},
  title        = {{XASP}: {eXplainable} 
                  Answer Set Programming},
  year         = {2023},
  howpublished = {\url{https://github.com/
                  alviano/xasp}},
  note         = {GitHub repository. 
                  Accessed: 2026}
}

@article{zhang2025benchmarking,
  author = {Zhang, Jian and 
            Wu, Peng and 
            London, Josh and 
            Tenney, David},
  title = {Benchmarking and Evaluating Large 
           Language Models in Phishing 
           Detection for Small and Midsize 
           Enterprises: A Comprehensive 
           Analysis},
  journal = {IEEE Access},
  volume = {13},
  pages = {28335--28352},
  year = {2025},
  doi = {10.1109/ACCESS.2025.3540075}
}

@inproceedings{lee2024multimodal,
  author = {Lee, Jehyun and 
            Lim, Peiyao and 
            Hooi, Bryan and 
            Divakaran, Dinil Mon},
  title = {Multimodal Large Language Models 
           for Phishing Webpage Detection 
           and Identification},
  booktitle = {2024 APWG Symposium on 
               Electronic Crime Research 
               (eCrime)},
  pages = {1--13},
  year = {2024},
  doi = {10.1109/eCrime66200.2024.00007}
}

@article{wang2019pdrcnn,
  title={PDRCNN: Precise phishing detection with recurrent convolutional neural networks},
  author={Wang, Weiping and Zhang, Feng and Luo, Xi and Zhang, Shigeng},
  journal={Security and Communication Networks},
  volume={2019},
  number={1},
  pages={2595794},
  year={2019},
  publisher={Wiley Online Library}
}

@article{uddin2025explainable,
  title={Explainable Machine Learning for Phishing Site Detection: A High-Efficiency Approach Using Boosting Models and SHAP},
  author={Uddin, Khandaker Mohammad Mohi and Biswas, Nitish and Rikta, Sarreha Tasmin and Nur-A-Alam, Md and Mostafiz, Rafid},
  journal={The Journal of Engineering},
  volume={2025},
  number={1},
  pages={e70110},
  year={2025},
  publisher={Wiley Online Library}
}

@article{barraclough2021intelligent,
  title={Intelligent cyber-phishing detection for online},
  author={Barraclough, Phoebe A and Fehringer, Gerhard and Woodward, John},
  journal={computers \& security},
  volume={104},
  pages={102123},
  year={2021},
  publisher={Elsevier}
}

@article{gupta2021novel,
  title={A novel approach for phishing URLs detection using lexical based machine learning in a real-time environment},
  author={Gupta, Brij B and Yadav, Krishna and Razzak, Imran and Psannis, Konstantinos and Castiglione, Arcangelo and Chang, Xiaojun},
  journal={Computer Communications},
  volume={175},
  pages={47--57},
  year={2021},
  publisher={Elsevier}
}

@article{ahammad2022phishing,
  title={Phishing URL detection using machine learning methods},
  author={Ahammad, SK Hasane and Kale, Sunil D and Upadhye, Gopal D and Pande, Sandeep Dwarkanath and Babu, E Venkatesh and Dhumane, Amol V and Bahadur, Mr Dilip Kumar Jang},
  journal={Advances in Engineering Software},
  volume={173},
  pages={103288},
  year={2022},
  publisher={Elsevier}
}

@article{le2018urlnet,
  title={URLNet: Learning a URL representation with deep learning for malicious URL detection},
  author={Le, Hung and Pham, Quang and Sahoo, Doyen and Hoi, Steven CH},
  journal={arXiv preprint arXiv:1802.03162},
  year={2018}
}

@article{li2025phishdebate,
  title={PhishDebate: an LLM-based multi-agent framework for phishing website detection},
  author={Li, Wenhao and Manickam, Selvakumar and Chong, Yung-wey and Karuppayah, Shankar},
  journal={arXiv preprint arXiv:2506.15656},
  year={2025}
}

@article{roy2024phishlang,
  title={PhishLang: A Real-Time, Fully Client-Side Phishing Detection Framework Using MobileBERT},
  author={Roy, Sayak Saha and Nilizadeh, Shirin},
  journal={arXiv preprint arXiv:2408.05667},
  year={2024}
}

@article{ejaz2023life,
  title={Life-long phishing attack detection using continual learning},
  author={Ejaz, Asif and Mian, Adnan Noor and Manzoor, Sanaullah},
  journal={Scientific reports},
  volume={13},
  number={1},
  pages={11488},
  year={2023},
  publisher={Nature Publishing Group UK London}
}

@article{aljofey2025bert,
  title={Bert-phishfinder: A robust model for accurate phishing url detection with optimized distilbert},
  author={Aljofey, Ali and Bello, Saifullahi Aminu and Lu, Jian and Xu, Chen},
  journal={IEEE Transactions on Dependable and Secure Computing},
  year={2025},
  publisher={IEEE}
}

@inproceedings{lokesh2026multimodal,
  title={Multimodal Phishing Website Detection using Hybrid Deep Learning and Ensemble Techniques},
  author={Lokesh, G and Ahamed, M Shaik Nabeel and Charan, Tandule and Khadri, Syed Rahamath Ali Sha and Rahul, S and Reddy, M Gopinath},
  booktitle={2026 6th International Conference on Expert Clouds and Applications (ICOECA)},
  pages={1536--1541},
  year={2026},
  organization={IEEE}
}

@inproceedings{bao2025phishsense,
  title={PhishSense: A LLM-Enhanced Multimodal Framework for Phishing Website Detection},
  author={Bao, Tiffany and Tang, Tingxuan},
  booktitle={2025 Silicon Valley Cybersecurity Conference (SVCC)},
  pages={1--3},
  year={2025},
  organization={IEEE}
}

@article{li2024concept,
  title={Concept drift adaptation by exploiting drift type},
  author={Li, Jinpeng and Yu, Hang and Zhang, Zhenyu and Luo, Xiangfeng and Xie, Shaorong},
  journal={ACM Transactions on Knowledge Discovery from Data},
  volume={18},
  number={4},
  pages={1--22},
  year={2024},
  publisher={ACM New York, NY}
}

@article{kalla2023phishing,
  title={Phishing website url’s detection using nlp and machine learning techniques},
  author={Kalla, Dinesh and Kuraku, Sivaraju},
  journal={Journal of Artificial Intelligence},
  volume={5},
  pages={145},
  year={2023},
  publisher={Tech Science Press}
}

@article{kocyigit2024enhanced,
  title={Enhanced feature selection using genetic algorithm for machine-learning-based phishing URL detection},
  author={Kocyigit, Emre and Korkmaz, Mehmet and Sahingoz, Ozgur Koray and Diri, Banu},
  journal={Applied sciences},
  volume={14},
  number={14},
  pages={6081},
  year={2024},
  publisher={MDPI}
}

@article{jalil2023highly,
  title={Highly accurate phishing URL detection based on machine learning},
  author={Jalil, Sajjad and Usman, Muhammad and Fong, Alvis},
  journal={Journal of Ambient Intelligence and Humanized Computing},
  volume={14},
  number={7},
  pages={9233--9251},
  year={2023},
  publisher={Springer}
}

@article{da2020heuristic,
  title={Heuristic-based strategy for Phishing prediction: A survey of URL-based approach},
  author={da Silva, Carlo Marcelo Revoredo and Feitosa, Eduardo Luzeiro and Garcia, Vinicius Cardoso},
  journal={Computers \& Security},
  volume={88},
  pages={101613},
  year={2020},
  publisher={Elsevier}
}

@article{mohammad2014intelligent,
  title={Intelligent rule based phishing websites classification},
  author={Mohammad, Rami and McCluskey, TL and Thabtah, Fadi},
  journal={IET Information Security},
  volume={8},
  number={3},
  pages={153--160},
  year={2014},
  publisher={Institution of Engineering and Technology}
}

@inproceedings{prakash2010phishnet,
  title={Phishnet: predictive blacklisting to detect phishing attacks},
  author={Prakash, Pawan and Kumar, Manish and Kompella, Ramana Rao and Gupta, Minaxi},
  booktitle={2010 Proceedings IEEE INFOCOM},
  pages={1--5},
  year={2010},
  organization={IEEE}
}

@mastersthesis{asif2023machine,
  title={Machine learning-based phishing detection using url features: A comprehensive review},
  author={Asif, Asif Uz Zaman},
  year={2023},
  school={Colorado State University}
}

@article{zieni2023phishing,
  title={Phishing or not phishing? A survey on the detection of phishing websites},
  author={Zieni, Rasha and Massari, Luisa and Calzarossa, Maria Carla},
  journal={IEEE Access},
  volume={11},
  pages={18499--18519},
  year={2023},
  publisher={IEEE}
}

@article{nappa2015malicia,
  title={The MALICIA dataset: identification and analysis of drive-by download operations},
  author={Nappa, Antonio and Rafique, M Zubair and Caballero, Juan},
  journal={International Journal of Information Security},
  volume={14},
  number={1},
  pages={15--33},
  year={2015},
  publisher={Springer}
}

@article{ban2023breaking,
  title={Breaking alert fatigue: AI-assisted SIEM framework for effective incident response},
  author={Ban, Tao and Takahashi, Takeshi and Ndichu, Samuel and Inoue, Daisuke},
  journal={Applied Sciences},
  volume={13},
  number={11},
  pages={6610},
  year={2023},
  publisher={MDPI}
}

@article{adane2023single,
  title={Single and hybrid-ensemble learning-based phishing website detection: Examining impacts of varied nature datasets and informative feature selection technique},
  author={Adane, Kibreab and Beyene, Berhanu and Abebe, Mohammed},
  journal={Digital Threats: Research and Practice},
  volume={4},
  number={3},
  pages={1--27},
  year={2023},
  publisher={ACM New York, NY}
}

@article{hannousse2021towards,
  title={Towards benchmark datasets for machine learning based website phishing detection: An experimental study},
  author={Hannousse, Abdelhakim and Yahiouche, Salima},
  journal={Engineering Applications of Artificial Intelligence},
  volume={104},
  pages={104347},
  year={2021},
  publisher={Elsevier}
}

@article{shafin2025explainable,
  title={An explainable feature selection framework for web phishing detection with machine learning},
  author={Shafin, Sakib Shahriar},
  journal={Data Science and Management},
  volume={8},
  number={2},
  pages={127--136},
  year={2025},
  publisher={Elsevier}
}

@article{do2022deep,
  title={Deep learning for phishing detection: Taxonomy, current challenges and future directions},
  author={Do, Nguyet Quang and Selamat, Ali and Krejcar, Ondrej and Herrera-Viedma, Enrique and Fujita, Hamido},
  journal={Ieee Access},
  volume={10},
  pages={36429--36463},
  year={2022},
  publisher={IEEE}
}

@inproceedings{odeh2021machine,
  title={Machine learningtechniquesfor detection of website phishing: A review for promises and challenges},
  author={Odeh, Ammar and Keshta, Ismail and Abdelfattah, Eman},
  booktitle={2021 IEEE 11th Annual Computing and Communication Workshop and Conference (CCWC)},
  pages={0813--0818},
  year={2021},
  organization={IEEE}
}

@article{das2019sok,
  title={SoK: a comprehensive reexamination of phishing research from the security perspective},
  author={Das, Avisha and Baki, Shahryar and El Aassal, Ayman and Verma, Rakesh and Dunbar, Arthur},
  journal={IEEE Communications Surveys \& Tutorials},
  volume={22},
  number={1},
  pages={671--708},
  year={2019},
  publisher={IEEE}
}

@article{GEBSER_KAMINSKI_KAUFMANN_SCHAUB_2019, 
title={Multi-shot ASP solving with clingo}, 
volume={19}, DOI={10.1017/S1471068418000054}, number={1}, journal={Theory and Practice of Logic Programming}, 
author={GEBSER, MARTIN and KAMINSKI, ROLAND and KAUFMANN, BENJAMIN and SCHAUB, TORSTEN}, year={2019}, pages={27–82}}

@book{brewka1997nonmonotonic,
  title={Nonmonotonic reasoning: an overview},
  author={Brewka, Gerhard and Dix, J{\"u}rgen and Konolige, Kurt and others},
  volume={73},
  year={1997},
  publisher={CSLI publications Stanford}
}

@article{putra2024analysis,
  title={Analysis of phishing attack trends, impacts and prevention methods: literature study},
  author={Putra, Fauzan Prasetyo Eka and Zulfikri, Achmad and Arifin, Goffal and Ilhamsyah, Revi Mario and others},
  journal={Brilliance: Research of Artificial Intelligence},
  volume={4},
  number={1},
  pages={413--421},
  year={2024}
}

\end{document}